\documentclass[manuscript, nonacm]{acmart}

\AtBeginDocument{%
  \providecommand\BibTeX{{%
    \normalfont B\kern-0.5em{\scshape i\kern-0.25em b}\kern-0.8em\TeX}}}

\usepackage{graphicx}
\usepackage{xcolor}
\usepackage{xspace}
\usepackage{amsmath}
\usepackage{bbm} 
\usepackage{subcaption}
\usepackage{algorithm}
\usepackage{algpseudocode}
\usepackage{mdwlist}
\usepackage{multirow} 
\usepackage{pifont}

\usepackage{wrapfig}
\definecolor{mygreen}{RGB}{0,160,100}
\definecolor{myorange}{RGB}{239,114,21}

\newcommand{\green}{\textcolor{mygreen}}

\newcommand{\kprune}{Kprune\xspace}

\newcommand{\va}{\boldsymbol{a}}
\newcommand{\vb}{\boldsymbol{b}}
\newcommand{\vc}{\boldsymbol{c}}
\newcommand{\vd}{\boldsymbol{d}}
\newcommand{\vg}{\boldsymbol{g}} 

\newcommand{\vw}{\boldsymbol{w}}
\newcommand{\vh}{\boldsymbol{h}}

\newcommand{\vx}{\boldsymbol{x}}
\newcommand{\vz}{\boldsymbol{z}}
\newcommand{\vzero}{\boldsymbol{0}}

\newcommand{\vzeta}{\boldsymbol{\zeta}}
\newcommand{\vxi}{\boldsymbol{\xi}}
\newcommand{\vbeta}{\boldsymbol{\beta}}
\newcommand{\vgamma}{\boldsymbol{\gamma}}

\newcommand{\mA}{\boldsymbol{A}}

\newcommand{\mW}{\boldsymbol{W}}
\newcommand{\mS}{\boldsymbol{S}}
\newcommand{\mU}{\boldsymbol{U}}
\newcommand{\mB}{\boldsymbol{B}}
\newcommand{\mC}{\boldsymbol{C}}
\newcommand{\mP}{\boldsymbol{P}}
\newcommand{\mE}{\boldsymbol{E}}
\newcommand{\mH}{\boldsymbol{H}} 
\newcommand{\mM}{\boldsymbol{M}}
\newcommand{\mX}{\boldsymbol{X}}
\newcommand{\mQ}{\boldsymbol{Q}}
\newcommand{\mK}{\boldsymbol{K}}
\newcommand{\mV}{\boldsymbol{V}}

\newcommand{\bR}{\mathbb{R}}
\newcommand{\bE}{\mathbb{E}}
\newcommand{\bD}{\mathbb{D}}

\newcommand{\cL}{\mathcal{L}}

\newcommand{\fisher}{\mathcal{I}}
\newcommand{\efisher}{\hat{\mathcal{I}}}
\newcommand{\cT}{\mathcal{T}}
\newcommand{\cS}{\mathcal{S}}

\newcommand{\llVert}{\left\lVert}
\newcommand{\rrVert}{\right\rVert}

\definecolor{ForestGreen}{rgb}{0.133, 0.545, 0.133}
\definecolor{torange}{rgb}{0.949, 0.522, 0.}
\definecolor{BrickRed}{rgb}{0.67, 0.29, 0.26}

\newcommand{\fcr}{\ding{171}} 
\newcommand{\dif}{\ding{58}} 
\newcommand{\comp}{\ding{64}} 

\begin{document}

\title{A Comprehensive Survey of Compression Algorithms for Language Models}

\author{Seungcheol Park}
\authornote{Equal contribution.}
\email{ant6si@snu.ac.kr}
\orcid{0009-0006-5315-7142}
\affiliation{%
  \institution{Seoul National University}
  \city{Seoul}
  \country{South Korea}
}
\author{Jaehyeon Choi}\authornotemark[1]
\email{jener05458@gmail.com}
\orcid{0000-0002-7214-5554}
\affiliation{%
  \institution{Seoul National University}
  \city{Seoul}
  \country{South Korea}
}
\author{Sojin Lee}\authornotemark[1]
\email{lsjlsj5846@snu.ac.kr}
\orcid{0009-0005-1884-6199}
\affiliation{%
  \institution{Seoul National University}
  \city{Seoul}
  \country{South Korea}
}
\author{U Kang}
\authornote{Corresponding author.}
\email{ukang@snu.ac.kr}
\orcid{0000-0002-8774-6950}
\affiliation{%
  \institution{Seoul National University}
  \city{Seoul}
  \country{South Korea}
}

\begin{abstract}
  
How can we compress language models without sacrificing accuracy?
The number of compression algorithms for language models is rapidly growing to benefit from remarkable advances of recent language models without side effects due to the gigantic size of language models, such as increased carbon emissions and expensive maintenance fees.
While numerous compression algorithms have shown remarkable progress in compressing language models, it ironically becomes challenging to capture emerging trends and identify the fundamental concepts underlying them due to the excessive number of algorithms.
In this paper, we survey and summarize diverse compression algorithms including pruning, quantization, knowledge distillation, low-rank approximation, parameter sharing, and efficient architecture design.
We not only summarize the overall trend of diverse compression algorithms but also select representative algorithms and provide in-depth analyses of them.
We discuss the value of each category of compression algorithms, and the desired properties of low-cost compression algorithms which have a significant impact due to the emergence of large language models.
Finally, we introduce promising future research topics based on our survey results.

\end{abstract}






\maketitle

\section{Introduction}
\label{sec:intro}
How can we compress language models without sacrificing accuracy?
The necessity of compressing language models~\cite{gpt3, llama, llama2, opt, bloom} has become significant since there are critical problems, such as increased carbon emissions and expensive maintenance fees, due to the gigantic size of language models despite their remarkable advancements.
A large number of compression algorithms including pruning~\cite{cofi, kpruning, pruner, auber, sparsegpt}, quantization~\cite{UI-BERT, ULLMint8, USmoothQuant, UOPTQ}, knowledge distillation (KD)~\cite{alpkd,mixkd,railkd,pkd}, low-rank approximation (LRA)~\cite{exploring, lora, mf, adalora}, parameter sharing~\cite{albert, gqa, subformer,peakd}, and those based on efficient neural architecture design~\cite{YCS21,nasbert, adabert, mobilebert} have been proposed, and they significantly improve the efficiency of language models;
however, an excessive number of research papers make it hard to figure out the overall trend and underlying fundamentals of compression algorithms.
Although existing surveys~\cite{SDLH20,SCMG20,SGCL21,SGKD21,SHAB21,SGA22,SKHW23} attempted to address this issue, their coverage is inadequate as they exclude low-cost compression algorithms which are 1) applicable to large language models (LLMs), and 2) currently prominent in the field of language model compression.

In this paper, we conduct an extensive survey of various compression algorithms including low-cost compression algorithms.
Our survey encompasses low-cost compression algorithms that are applicable to LLMs considering their prominence, and elaborates on the transition from high-cost compression algorithms to low-cost ones.
We compare the performance of compression algorithms and summarize the overall trend of their development.
We select representative algorithms that significantly impact language model compression, and conduct in-depth analyses of these algorithms.
We discuss the contribution of each field of compression algorithms, and introduce desired properties for successful low-cost compression algorithms for LLMs.
Finally, we propose promising research topics based on the discussion.

We summarize the major contribution of our paper as follows.
\begin{itemize*}
    \item \textbf{Overview.}
    We summarize an overall trend of compression algorithms for language models.
    Our paper covers diverse types of compression algorithms for both high-cost and low-cost compression algorithms.
    \item \textbf{Analysis.}
    We select three representative algorithms each from pruning, quantization, and other compression algorithms, elaborating their details to provide meaningful insights.
    \item \textbf{Discussion.}
    We provide valuable discussion regarding compression algorithms\green{,} and introduce promising future research topics.
\end{itemize*}

The rest of this paper is organized as follows.
In Section~\ref{sec:prelim}, we formally define a language model compression problem and describe preliminaries of algorithms to solve the language model compression problem.
In Sections~\ref{sec:pruning}-\ref{sec:others}, we summarize existing compression algorithms and provide a detailed explanation of the representative algorithms regarding pruning, quantization, and other compression algorithms.
We discuss the current compression algorithms and promising future research areas in Section~\ref{sec:disc} and conclude in Section~\ref{sec:conclusion}.

\section{Preliminaries}
\label{sec:prelim}
In this section, we formally define the pretrained language model (PLM) compression problem in Section~\ref{subsec:prelim1},
introduce a Transformer architecture which dominates the state-of-the-art language models in Section~\ref{subsec:prelim2},
overview PLM compression algorithms in Section~\ref{subsec:prelim3},
and describe the backgrounds to solve the PLM compression problem in Section~\ref{subsec:prelim4}.
Table~\ref{tab:symbol} presents notations frequently used in this paper.

\begin{table}[]
	\small
	\caption{Symbols and descriptions.}
	\label{tab:symbol}
		\begin{tabular}{@{}cl|cl@{}}
			\toprule
			\textbf{Symbol} & \multicolumn{1}{c}{\textbf{Description}} & \textbf{Symbol} & \multicolumn{1}{c}{\textbf{Description}} \\ \midrule
			$\mathcal{T}$& a pre-trained language model (PLM) &
			$\mathbb{D}$, $\mathbb{S}$ & target and calibration datasets \\
			$\mathcal{S}$ & a compressed language model &
			$(\vd_i,y_i)$ & a tuple of the $i$th instance and its label\\
			\midrule
			$M(\cdot)$ & a multi-head attention (MHA) function&
			$\mX$ & an input matrix of a sublayer \\
			$F(\cdot)$ & a feedforward network (FFN) function &
			$\vx_i$ & the $i$th token embedding in $\mX$ \\
			$\mA_i$ & an output of the $i$th attention head &
			$d$ & the dimension of token embeddings \\
			$\mQ_i$, $\mK_i$, $\mV_i$ & query, key, and value matrices of $i$th attention head &
			$H$ & the number of attention heads \\
			$\mW^Q$, $\mW^K$, $\mW^V$, $\mW^O$ & weights in an MHA sublayer &
			$a(\cdot)$ & a position-wise affine transformation \\
			$\vb^Q$, $\vb^K$, $\vb^V$, $\vb^O$ & biases in an MHA sublayer &
			$S(\cdot)$ & a column-wise softmax function \\
			$\mU^{I}$, $\mU^{O}$ & weights in an FFN sublayer &
			$\mathcal{N}(\cdot)$ & a layer normalization function \\
			$\vc^{I}$, $\vc^{O}$ & biases in an FFN sublayer &
			$||$ & vertical concatenation \\
			\midrule
			$\cL$ & a task-specific object function (cross-entropy loss) &
			$\vg$ & a gradient of $\cL$ \\
			%
			$\mH$ & a Hessian matrix of $\cL$ &
			$Q(\cdot)$ & a quantization operator (quantizer) \\ \bottomrule
	\end{tabular}
\end{table}

\subsection{Pretrained Language Model Compression Problem}
\label{subsec:prelim1}
The pretrained language model (PLM) compression problem is defined as follows.
We have an accurate and large PLM $\mathcal{T}$ and dataset $\mathbb{D}=\{(\vd_i,y_i)\}$ where $(\vd_i,y_i)$ is the tuple of the $i$th data point and its label.
Our goal is to generate a tiny but still accurate compressed language model $\mathcal{S}$ that meets cost constraints, usually, in terms of FLOPs~\cite{KKMHKG22,kcm}, memory usage~\cite{exploring, albert}, or latency~\cite{cofi,bmp} upper bounds.
In this paper, we express the reduced memory usage ratio compared to the original PLM as the compression rate.
If such information is not reported, we use the reduction in FLOPs as the compression rate and annotate with \fcr.
We assume that we have an accurate PLM, according to the practical settings in a dominant number of existing works~\cite{cofi,ULLMint8,USmoothQuant,exploring,lora} that utilize sophisticated PLMs~\cite{gpt3, opt, llama, llama2} trained on an immense corpus.
The label $y$ can be omitted for the unlabeled tasks such as language modeling~\cite{pruner,ULLMint8,USmoothQuant}.
For low-cost compression algorithms~\cite{KKMHKG22,UZeroQuant,UZeroQuant2,kcm}, we use a small calibration dataset $\mathbb{S} \subset\mathbb{D}$ to minimize their cost.

\subsection{Transformer Architecture}
\label{subsec:prelim2}
\begin{figure}[t]
	\minipage{0.38\textwidth}
	\includegraphics[width=\linewidth]{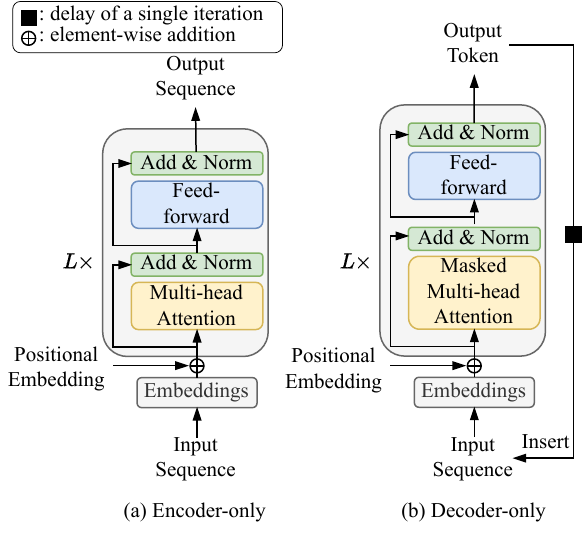}
	\caption{Two variants of Transformer architecture.}\label{fig:2_models}
	\endminipage
	\hfill	
	\minipage{0.55\textwidth}
	\includegraphics[width=\linewidth]{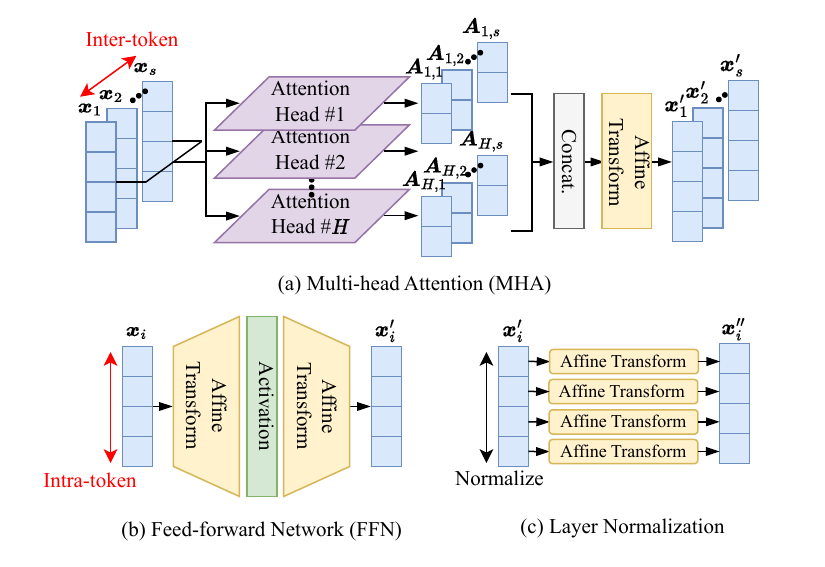}
	\caption{Illustrations of multi-head attention (MHA), feed-forward network (FFN), and layer normalization.}\label{fig:2_layers}
	\endminipage
\end{figure}

We elaborate on Transformer architecture which is dominantly used for PLMs~\cite{gpt3, opt, llama, llama2} in natural language processing (NLP).
Transformer~\cite{transformer} is a language model that refines token embeddings reflecting the contextual information through a self-attention mechanism~\cite{CDL16, PTD16, transformer} to utilize them in downstream tasks.
Transformer is composed of multiple layers consisting of multi-head attention (MHA) and feed-forward network (FFN) sublayers.
MHA sublayers capture the contextual inter-token information via a self-attention mechanism and FFN sublayers polish intra-token information with a position-wise feedforward network.
Original Transformer architecture~\cite{transformer} comprises Encoder and Decoder where Encoder encodes input token embeddings considering contextual information and Decoder generates output tokens by decoding the encoding results.
However, we use encoder-only and decoder-only Transformers according to downstream tasks based on the previous works demonstrating the impressive performance of encoder-only~\cite{bert,roberta,albert,distilbert,electra} and decoder-only~\cite{gpt3,opt,llama,bloom} PLMs.
We illustrate encoder-only and decoder-only Transformer architectures in Fig.~\ref{fig:2_models}.
Encoder-only Transformers (a)~\cite{bert,roberta,albert,distilbert,electra} generate useful embeddings that reflect contextual information within an input sequence via (bidirectional) self-attention.
Encoder-only Transformers are used for natural language understanding (NLU) tasks including sentence similarity~\cite{mrpc,stsb}, natural language inference~\cite{mnli,qnli}, and question answering~\cite{squad,squadv2}.
On the other hand, decoder-only Transformers (b)~\cite{gpt3,opt,llama,bloom} autoregressively predict output tokens via (unidirectional) masked self-attention which attends only current and previous tokens.
We concatenate an output token to the end of an input sequence, and feed the augmented input sequence to the next iteration.
Decoder-only models are used for natural language generation (NLG) tasks, such as question answering~\cite{naturalqs,triviaqa}, mathematical reasoning~\cite{math,gsm8k}, and code generation~\cite{humaneval,mbpp}.
Note that decoder-only Transformers are broadly applicable since text-based language problems are convertible into a text-to-text format~\cite{t5}.

Fig.~\ref{fig:2_layers} illustrates the three mechanisms used in Transformer.
Assume that an input $\mX=[\vx_1;\vx_2;...;\vx_s]\in\mathbb{R}^{d\times s}$ consists of $d$-dimensional $s$ token embeddings from the previous layer.
The output of each sublayer is computed as either $\mathcal{N}(\mX + M(\mX))$ for MHA sublayers or $\mathcal{N}(\mX + F(\mX))$ for FFN sublayers, where $M(\mX)$ and $F(\mX)$ are MHA and FFN functions, respectively.
$\mathcal{N}(\cdot)$ represents layer normalization~\cite{layernorm}.
MHA captures contextual information across tokens via self-attention.
We utilize multiple attention heads to capture diverse patterns on attention maps and use shrunk dimension $d_h=d/H$ to reduce the cost of using multiple heads where $H$ is the number of attention heads.
We obtain an output $M(\mX)$ by concatenating the output of $H$ attention heads and applying a position-wise affine transformation as in Equation~\eqref{eq:2_mha}.
$||$ represents vertical concatenation along embedding dimension and $\mA_i \in \mathbb{R}^{d_h \times s}$ is an output of the $i$th attention head.
$a(\;\cdot\;; \mW^O, \vb^O):\mathbb{R}^{d\times s} \to \mathbb{R}^{d\times s}$ represents a position-wise affine transformation parameterized by weight $\mW^O\in\mathbb{R}^{d\times d}$ and bias $\vb^{O}\in\mathbb{R}^{d}$.
We compute the $j$th column of $a(\;\cdot\;; \mW^O, \vb^O)$ as the affine transformation of the $j$th token embedding as in Equation~\eqref{eq:2_affine}.
We compute an output $A_i$ of attention heads via scaled dot-product attention~\cite{transformer}.
We compute query $\mQ_i$, key $\mK_i$ and value $\mV_i$ matrices via position-wise affine transformations of the input $\mX$ with different weights $\mW_i^Q$, $\mW_i^K$, $\mW_i^V\in\mathbb{R}^{d_h\times d}$ and biases $\vb_i^Q$, $\vb_i^K$, $\vb_i^V\in\mathbb{R}^{d_h}$ in each attention head.
Note that $\mQ_i, \mK_i, \mV_i\in\mathbb{R}^{d_h\times s}$, and $S(\cdot)$ is a column-wise softmax function across tokens for scoring.
\begin{align}\label{eq:2_mha}
	M(\mX)  = a \left(||_{i=1}^{H}  \mA_i ;\mW^O, \vb^O \right)
	\;\; whe&re \;\;
	\mA_i = \mV_i S \left( \left(\mK_i^T \mQ_i \right) / \sqrt{d_h} \right) \\
	\label{eq:2_affine}
	a\left(||_{i=1}^{H}  \mA_i; \mW^O, \vb^O \right)_{:,j} &=\mW^O \left(||_{i=1}^{H}  \mA_i\right)_{:,j} + \vb^O
\end{align}
FFN separately and identically refines each token embedding through two position-wise affine transformations with weights $\mU^{I}\in \mathbb{R}^{d_f \times d}$, $\mU^{O}\in \mathbb{R}^{d \times d_f}$ and biases $\vc^{I}\in \mathbb{R}^{d_f}$, $\vc^{O}\in \mathbb{R}^{d}$.
We transform each token embedding into the higher dimension of $d_f$ to capture useful patterns within a token embedding; conventionally, we use $d_f=4d$.
$\sigma$ is a non-linear activation function, such as ReLU~\cite{relu} or GELU~\cite{gelu}.
\begin{equation}\label{eq:2_ffn}
F(\mX) = a\left( \sigma\left( a\left(
				\mX; \mU^{I}, \vc^{I}
				\right) \right); \mU^{O}, \vc^{O}\right)
\end{equation}

We describe an example of layer normalization~\cite{layernorm} in Fig.~\ref{fig:2_layers} (c).
Layer normalization is a modification of batch normalization~\cite{batchnorm} adaptable to flexible lengths of sequences to be appropriate for NLP.
Layer normalization normalizes each token embedding and performs element-wise affine transformations.
The normalized output of $j$th element of the $i$th input token $\vx_i'$ is described in Equation~\eqref{eq:2_ln} where $\mu_i=\frac{1}{d}\sum_{j=1}^{d}\left(x'_i\right)_j$ and $\nu_i=\frac{1}{d}\sum_{j=1}^{d} \left( \left( x'_i \right)_j-\mu \right)^2$ are the mean and variance of the elements in $\vx_i'$.
$\vbeta \in \mathbb{R}^{d}$ and $\vgamma \in \mathbb{R}^d$ are learnable parameters for the element-wise affine transformations.
\begin{equation}\label{eq:2_ln}
	(x_{i}'')_j = \mathcal{N}(x'_i)_j = \frac{(x_{i}')_j-\mu_i}{\sqrt{\nu_i}} \gamma_j + \beta_j
\end{equation}

\subsection{Pretrained Language Model Compression Algorithms}
\label{subsec:prelim3}

The goal of pretrained language model (PLM) compression is to reduce the cost of PLMs; we reduce the cost of representing, and computing weights and intermediate activations which take the majority portion of the cost of PLMs.
We categorize PLM compression algorithms based on how we reduce the cost of weights and activations as follows.
\begin{itemize*}
	\item\textbf{Pruning (Section~\ref{sec:pruning}).}
	Identifying and removing unnecessary weights of PLMs.
	The size of activation is also decreased when we use large group of parameters as a pruning granularity.
	\item\textbf{Quantization (Section~\ref{sec:quantization}).}
	Reducing the bit-length for representing weights.
	We achieve acceleration in matrix multiplications when we reduce the bit-length of the activations as well.
	\item\textbf{Knowledge distillation (Section~\ref{subsec:o2}).}
	Improving the accuracy of compressed models by transferring useful knowledge of the PLM.
	We combine knowledge distillation (KD) with other compression algorithms to improve the accuracy of the compressed models.
	\item\textbf{Low-rank approximation (Section~\ref{subsec:o3}).}
	Reducing the cost of PLMs by substituting the weights with the multiplication of low-dimensional matrices based on the low-rank assumption.
	\item\textbf{Parameter sharing (Section~\ref{subsec:o4}).}
	Sharing weights in different modules in PLMs.
	It is crucial to find a pair of weights that do not decrease accuracy after the sharing.
	\item\textbf{Efficient architecture design (Section~\ref{subsec:o5}).}
	Designing a cost-efficient architecture that requires low costs for memorization and computation.
\end{itemize*}
We additionally classify PLM compression algorithms into low-cost and high-cost ones considering the importance of compressing massive large language models (LLMs), independently of the aforementioned classification criterion.
High-cost compression algorithms require extensive retraining processes on large datasets.
They consequently generate more accurately compressed models but require intractable time to compress LLMs.
On the other hand, low-cost compression algorithms perform only a simple weight-tuning process on small calibration data.
Low-cost compression algorithms are applicable to LLMs; however, they suffer from severe accuracy degradation, and improving the accuracy is their main concern.

\subsection{Backgrounds}
\label{subsec:prelim4}
In this section, we elaborate on fundamental backgrounds to solve PLM compression problems.
\subsubsection{Taylor Expansion}

PLM compression is a procedure that imposes perturbations $\delta \vw$ on weights $\vw$ in a PLM.
Examples include setting weights to zeros~\cite{obd, obs}, low-precision representations~\cite{UOPTQ,UI-BERT,USensiMix}, and low-rank representations~\cite{exploring,fwsvd}.
Taylor expansion is useful to estimate the amount $\delta \cL$ of the increment in an objective function $\cL$ after applying the perturbation $\delta \vw$ on weights as in Equation~\eqref{eq:2_taylor}~\cite{obd}.
\begin{equation} \label{eq:2_taylor}
\delta \cL = \delta \vw^T \vg + \frac{1}{2} \delta \vw ^T \mH \delta \vw + \mathcal{O}(\delta \vw^3)
\approx \frac{1}{2} \delta \vw ^T \mH \delta \vw
\end{equation}
$\vg= \bE \left[ \frac{\partial \cL (\vw^*)}{\partial \vw} \right]$ is the gradient and $\mH= \bE \left[\frac{\partial^2 \cL (\vw^*)}{\partial \vw^2} \right] $ is the Hessian of $\cL$ with respect to $\vw$.
We approximate $\vg \approx 0$ since the PLM is pretrained at a local minimum whose gradient is zero.
We neglect $\mathcal{O}(\delta \vw^3)$ since the size of the perturbation is sufficiently small.
As a result, we have only a quadratic term with the Hessian matrix $\mH$. 
Using Taylor expansion is efficient since we compute $\delta \cL$ for any $\delta \vw$ by computing the Hessian matrix once;
we do not have to manually compute $\delta \cL$ for each $\delta \vw$.
We utilize the derivation in Equation~\eqref{eq:2_taylor} in various compression algorithms, for example, measuring sensitivity in quantization~\cite{NUSqueezeLLM}, measuring saliency score in pruning~\cite{obd,obs,dynabert,KKMHKG22}, and selecting rank-1 components in low-rank approximation~\cite{fwsvd}.
\subsubsection{Lagrangian}
Lagrangian function is
used to solve a constrained optimization problem.
Consider the problem of minimizing an objective function $f(\vw)$ with equality constraints $\va^T\vw + \vb =\vzero$ on weights $\vw$.
In PLM compression problems, the objective function $f(\vw)$ is either cross-entropy loss or layer-wise least-square loss.
$\va$ and $\vb$ are vectors of constants to state constraints; e.g.,
$a_q = 1$ and $b_q=0$ represent that the $q$th weight is pruned. 
The solution of the problem is computed by first forming the following Lagrangian function
\begin{equation}\label{eq:2_dual}
L(\vw, \lambda) = f(\vw) + \lambda (\va^T\vw + \vb)
\end{equation}
%
and finding the stationary point of $L(\vw, \lambda)$ with respect to $\vw$;
i.e., $\frac{\partial L(\vw, \lambda)}{\partial \vw}=\vzero$.
For example, optimal brain surgeon (OBS)~\cite{obs} finds the optimal perturbations (surgery) on weights through the Lagrangian function.
The optimal surgery in OBS is crucial and widely used in state-of-the-art compression algorithms both in pruning~\cite{sparsegpt,ziplm} and quantization~\cite{UOPTQ,URPTQ,UAWQ}.
We elaborate on more details about these algorithms in Sections~\ref{subsec:p4} and~\ref{subsec:Qalg}.
%
\subsubsection{Fisher Information Matrix}
\label{subsubsec:fisher}
A Fisher information matrix is an information measure in statistics.
However, in model compression, we use a Fisher information matrix to approximate a Hessian matrix since the Hessian matrix $\mH$ of the cross-entropy loss $\cL$ is able to be approximated by the Fisher information matrix $\fisher$ (see Lee et al.~\cite{HessianFisher} for derivation).
As in Equation~\eqref{eq:2_fisher}, we approximate the Fisher information matrix $\fisher$ as an empirical Fisher information matrix $\efisher$ estimated on the sample dataset $\bD$ since computing the exact form is intractable.
$\vd$ is a data point in $\bD$.
\label{key}
\begin{equation} \label{eq:2_fisher}
\bE_{\vd}[\mH] = \bE_{\vd} \left[ \frac{\partial^2 \cL(\vd;\vw)}{\partial \vw^2} \right]
\approx
\frac{1}{|\bD|} \sum_{\vd\in\bD} \left[ \left(  \frac{\partial \cL(\vd;\vw)}{\partial \vw} \right)
\left(  \frac{\partial \cL(\vd;\vw)}{\partial \vw} \right)^T \right]
=
\efisher
\end{equation}
We use the Fisher information matrix to approximate the Hessian in Equation~\eqref{eq:2_taylor} since computing an exact Hessian matrix of cross-entropy loss is computationally expensive.
Note that Equation~\eqref{eq:2_fisher} holds only when the objective function is a cross-entropy loss; for other objective function, we need to find alternative ways to compute the Hessian. 
For example, when our loss function is a layer-wise least-square loss~\cite{obs,sparsegpt,UOPTQ}, we directly compute the Hessian matrix. 

\subsubsection{Straight-Through Estimator (STE)}
\begin{figure}[t]
	\centering
	\includegraphics[width=\textwidth]{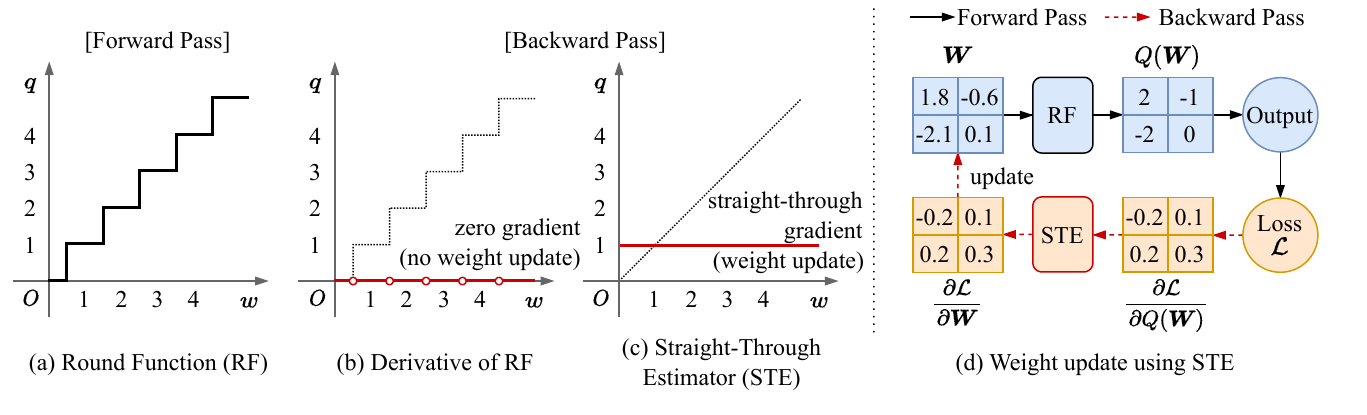}
	\caption{
		Illustration of how Straight-Through Estimator works.
		}
	\label{fig:STE}
\end{figure}
STE~\cite{STE} is a technique of passing upstream gradients directly to the next node during back propagation.
We leverage STE to train a model that entails non-differentiable functions, such as a round function (RF) in Fig.~\ref{fig:STE} (a).
Weights cannot be updated in the backward pass of RF, as depicted in Fig.~\ref{fig:STE} (b), since the gradient of RF is either 0 or not defined.
We therefore substitute the derivative of RF with STE (Fig.~\ref{fig:STE} (c)), which enables the gradient update by having gradients of 1 in all regions.
Fig.~\ref{fig:STE} (d) shows an example of a weight update using STE.
We quantize a weight matrix $\mW$ using a RF $Q(\cdot)$ and compute the loss $\cL$ in the forward pass.
In the backward pass, the gradient of $\mW$ with respect to the loss $\cL$ is approximated via STE as in Equation~\eqref{eq:2_ste}.
\begin{equation}\label{eq:2_ste}
	\frac{\partial\cL}{\partial\mW}=
	\frac{\partial\cL}{\partial Q(\mW)}\cdot\frac{\partial Q(\mW)}{\partial\mW}\approx\frac{\partial\cL}{\partial Q(\mW)}.
\end{equation}
%
STE is one of the core methodologies for quantization schemes that involve weight update with gradients, named quantization-aware training (QAT).
%
\subsubsection{Singular Value Decomposition (SVD)}
Singular value decomposition (SVD) is a factorization of a matrix.
SVD reduces the number of weights to represent weight matrices in pretrained language models (PLMs).
Given a weight matrix $\mW\in\mathbb{R}^{m\times n}$, we decompose the matrix with SVD as in Equation~\eqref{eq:2_svd}.
$\mU\in\mathbb{R}^{m\times r}$ and $\mV\in\mathbb{R}^{n\times r}$ are column orthogonal matrices where $r$ is the rank of $\mW$.
$\mS$ is a diagonal matrix consisting of $r$ singular values $\{s_i\}_{i=1}^r$.
\begin{equation}\label{eq:2_svd}
\mW = \mU \mS \mV^T = \sum_{i=1}^{r}s_i \mU_{:,i} (\mV_{:,i})^T
\approx \sum_{i=1}^{k}s_i \mU_{:,i} (\mV_{:,i})^T
=\overbrace{\left( \mU_{:,:k} (\mS_{:k})^{1/2} \right)}^{m\times k}
 \overbrace{\left( (\mS_{:k})^{1/2} (\mV_{:,:k})^T \right)}^{k\times n}
 \;\; k\ll r
\end{equation}
We transform the equation into the sum of rank-1 components and approximate $\mW$ as the sum of the $k$ rank-1 matrices with the $k$ largest singular values since a small number of singular values dominates other singular values in weight matrices in language models~\cite{fwsvd, collabo}.
We decompose $\mS$, and integrate it with the orthogonal matrices to remove redundant weights.
As a result, the number of parameters for representing $\mW$ is reduced from $mn$ to $(m+n)k$; however, SVD results in a computational overhead due to the increased number of matrix multiplications.

\section{Pruning}
\label{sec:pruning}
In this section, we introduce pruning algorithms, which identify and prune redundant components in PLMs to generate compact and accurate language models.
We give an overview of pruning algorithms in Section~\ref{subsec:p1}, and elaborate on pruning granularities and pruning strategies in Sections~\ref{subsec:p2} and~\ref{subsec:p3}, respectively.
In Section~\ref{subsec:p4}, we delve into SparseGPT~\cite{sparsegpt} which is the state-of-the-art pruning algorithm for massive-scale LLMs up to 175B.

\subsection{Overview}
\label{subsec:p1}

Pruning is a compression algorithm that eliminates the superfluous components in neural networks.
Pruning algorithms~\cite{kpruning,ziplm,sparsegpt,pruner} have demonstrated the ability to compress PLMs to small models whose inference is much faster than PLMs on commodity hardwares.
For instance, ZipLM~\cite{ziplm} accelerates the inference speed of BERT~\cite{bert} up to 15 times with a small accuracy degradation.
%
\begin{figure}[t]
	\includegraphics[width=\linewidth]{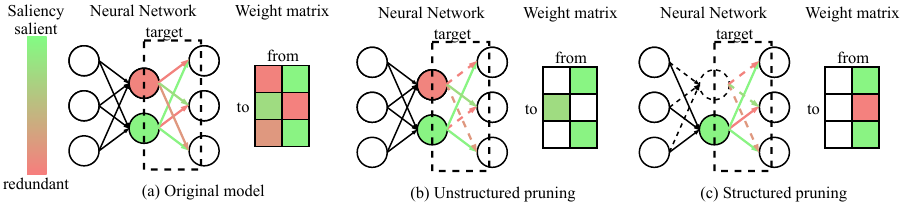}
	\caption{An example of pruning of a neural network with two layers using unstructured pruning or structured pruning.}\label{fig:3_pruning}
	\vspace{-0.3cm}
\end{figure}
In Fig.~\ref{fig:3_pruning}, we illustrate an example of pruning a three-layer neural network (a), specifically targeting the weights in the second layer.
The weight matrix corresponding to the pruning targets is shown on the right side of the neural network.
We first measure the saliency of each weight (arrow) and neuron (circle) to identify salient weights and neurons to be preserved.
We color salient components in green and redundant components in red.
After measuring the saliency, we have to choose the granularity of the pruning, which is either a weight or a neuron.
(b) represents an unstructured pruning algorithm whose pruning granularity is a weight.
We get an accurate and compact model that has only salient green-colored weights.
However, the pruned weight matrix is sparse and requires sparse matrix representation, such as CSR or COO formats, which are inefficient to store and compute.
(c) represents a structured pruning algorithm which prunes neurons, and generates a dense weight matrix that is efficient to store and compute.
In general, the compressed models pruned with larger granularities (e.g., neurons) are more efficient than those with smaller granularities (e.g. weights); on the other hand, large granularities lead to inaccuracy.
We describe more details of the accuracy-efficiency trade-off regarding pruning granularity in Section~\ref{subsec:p2}.

To maximize the accuracy of pruned models, it is essential to carefully design a strategy to minimize pruning errors which refer to the distortion of the model's output induced by pruning components in PLMs.
For example, we need to update the remaining weights (solid lines) in Fig.~\ref{fig:3_pruning} to compensate for the pruning errors induced by the pruning of redundant weights (dotted lines).
It is crucial for pruning strategies to consider the following two problems to achieve high accuracy: 1) how to choose components to prune and 2) how to adequately compensate for pruning errors.
We name the first issue of finding suitable pruning masks as a mask search problem, and the second issue of reducing pruning errors as an error compensation problem.
We divide pruning strategies into high-cost and low-cost algorithms based on their computational costs.
High-cost pruning algorithms jointly identify and remove unnecessary components by time-consuming retraining processes on a large dataset, compensating for pruning errors by tuning the weights in unpruned components.
Although high-cost pruning algorithms generate accurate and compact models, they are too expensive to be used for LLMs which have more than billions of parameters.
Low-cost pruning algorithms significantly reduce the computational cost of pruning by dividing pruning processes into efficient mask search and low-cost error compensation on a small calibration dataset.
For instance, \kprune~\cite{kpruning} prunes BERT within 10 minutes on the MNLI dataset while high-cost pruning algorithms~\cite{dynabert} take up to 40 hours on the same dataset. 
However, low-cost algorithms suffer from severe accuracy degradation in high-compression regimes.
We describe more details about pruning strategies and pruning costs in Section~\ref{subsec:p3}.

We summarize pruning algorithms for encoder-only Transformers in Table~\ref{tab:3_eo} and decoder-only Transformers in Table~\ref{tab:3_do}.
We represent the pruning granularity and cost of each pruning algorithm, and group them accordingly.
We report the accuracy difference before and after pruning, and report speedup if available.

\begin{figure}[t]
	\includegraphics[width=0.7\linewidth]{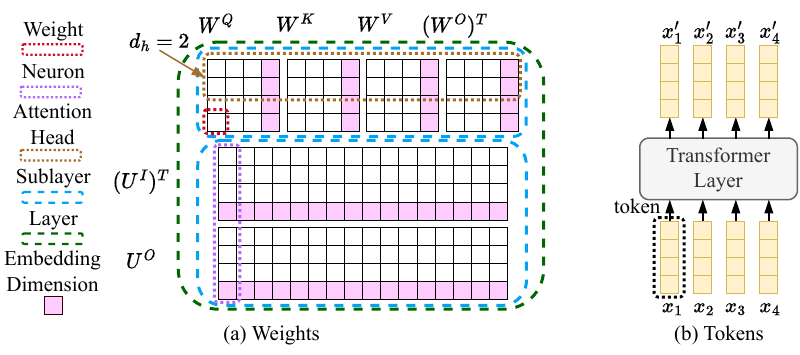}
	\caption{
		Illustration of diverse pruning granularities regarding weights (a) and tokens (b).
		Dotted square boxes indicate each pruning granularity and we color the weights for the granularity of the embedding dimension in pink for simplicity.
		$d_h$ represents the dimension of token embeddings in attention heads.
	}\label{fig:3_granularity}
\end{figure}

\setlength{\tabcolsep}{5pt}
\begin{table}[t]
	\caption{
			Comparison of pruning algorithms for encoder-only Transformers on MNLI, QQP, and $\text{SQuAD}_{1.1}$ (SQD) benchmarks.}
	\label{tab:3_eo}
	\begin{tabular}{@{}lccccrrrrc@{}}
		\toprule
		\multicolumn{1}{c}{\multirow{2}{*}{\textbf{Method}}} & \multirow{2}{*}{\textbf{PLM$^{\text{\ding{168}}}$}} & \multirow{2}{*}{\textbf{\begin{tabular}[c]{@{}c@{}}Pruning\\ Granularity$^\S$\end{tabular}}} & \multirow{2}{*}{\textbf{\begin{tabular}[c]{@{}c@{}}Pruning\\ Cost\end{tabular}}} & \multirow{2}{*}{\textbf{\begin{tabular}[c]{@{}c@{}}Comp.\\ Rate\end{tabular}}} & \multicolumn{4}{c}{\textbf{Accuracy/F1 Difference$^\text{\dif}$ (\%p)}}  &
		\multicolumn{1}{c}{\multirow{2}{*}{\textbf{Speedup}}} \\
		\multicolumn{1}{c}{} &&  & &  & \multicolumn{1}{c}{\textbf{MNLI}} & \multicolumn{1}{c}{\textbf{QQP}} & \multicolumn{1}{c}{\textbf{SQD}} & \multicolumn{1}{c}{\textbf{Avg.}} & \multicolumn{1}{c}{}  \\ \midrule
		MvP~\cite{mvp}& Bb & W& High & 90.0\%  & -3.30 & -1.20& -3.20& -2.57 & \multicolumn{1}{c}{-} \\
		LEAP~\cite{leap}& Bb & W& High & 90.0\%  & -3.30 & -0.20& -4.80& -2.77 & \multicolumn{1}{c}{-} \\
		CAP~\cite{cap}& Bb & W& High & 90.0\%  & -2.50 & -0.20& -2.80& -1.83 & \multicolumn{1}{c}{-} \\
		oBERT~\cite{obert}   & Bb & W& High & 90.0\%  & -1.34 & -0.17& -0.56& -0.69 & \multicolumn{1}{c}{-} \\
		oBERT~\cite{obert}   & Bb & W& High & 97.0\%  & -3.54 & -0.83& -3.89& -2.75 & \multicolumn{1}{c}{-} \\
		\midrule
		DynaBERT~\cite{dynabert} & Bb & N, H & High & 75.0\%$^{\text{\fcr}}$ & -0.90 & -0.20& -0.60& -0.57 & \multicolumn{1}{c}{-} \\
		DynaBERT~\cite{dynabert}$^{\text{\ding{169}}}$& Bb & N, H & High & 75.0\%$^{\text{\fcr}}$ & -4.30 & \multicolumn{1}{c}{-}& \multicolumn{1}{c}{-}  & -4.30 & 3.20$\times$\\
		SNIP~\cite{snip}& Bb & N, H, S & High & 75.0\%  & -6.40 & -2.80& \multicolumn{1}{c}{-}  & -4.60 & \multicolumn{1}{c}{-} \\
		EBERT~\cite{ebert}   & Bb & N, H & High & 50.0\%$^{\text{\fcr}}$ & -2.30 & -0.50& \multicolumn{1}{l}{}   & -1.40 & \multicolumn{1}{c}{-} \\
		BMP\cite{bmp} & Bb & N, H & High & 75.2\%  & \multicolumn{1}{c}{-} & \multicolumn{1}{c}{-}& -0.80& -0.80 & 2.25$\times$\\
		CoFi~\cite{cofi}& Bb & E, N, H, S& High & 75.0\%$^{\text{\fcr}}$ & -0.05 & \multicolumn{1}{c}{-}& -0.41& -0.23 & 3.08$\times$\\
		CoFi~\cite{cofi}& Bb & E, N, H, S& High & 90.0\%$^{\text{\fcr}}$ & -2.19 & \multicolumn{1}{c}{-}& -4.32& -3.26 & 6.68$\times$\\
		CAP~\cite{cap}& Bb & N, H & High & 90.0\%  & -3.50 & -0.70& -8.10& -4.10 & \multicolumn{1}{c}{-} \\
		Sajjad et al.~\cite{sajjad23}   & Bb & L& High & 50.0\%  & -2.91 & -0.72& \multicolumn{1}{c}{-}  & -1.82 & 2.00$\times$\\
		ZipLM~\cite{ziplm}   & Bb & N, H, S & High & -& -1.50 & -0.20& -2.50& -1.40 & 7.00$\times$\\
		ZipLM~\cite{ziplm}   & Bb & N, H, S & High & -& -3.90 & -0.80& -7.20& -3.97 & 15.00$\times$\\
		\midrule
		PoWER-BERT~\cite{powerbert}& Bb & T& High & -& -0.80 & -1.00& \multicolumn{1}{c}{-}  & -0.90 & 3.42$\times$\\
		LTP~\cite{ltp}& Rb & T& High & 50.5\%$^{\text{\fcr}}$ & -1.00 & -0.70& \multicolumn{1}{c}{-}  & -0.85 & 1.99$\times$\\
		\midrule
		Kwon et al.~\cite{KKMHKG22}$^\dag$& Bb & N, H & Low& 75.0\%$^{\text{\fcr}}$ & -21.67& -19.42 & -48.56   & -29.88& 1.75$\times$$^{\ddag\dag}$ \\
		KCM~\cite{kcm}& Bb & N& Low& 20.0\%$^{\text{\fcr}}$ & -7.29 & -1.85& -7.19& -5.44 & 1.21$\times$$^{\ddag\dag}$ \\
		\kprune~\cite{kpruning}& Bb & N, H & Low& 75.0\%$^{\text{\fcr}}$ & -5.18 & -3.71& -9.26& -6.05 & 2.08$\times$$^{\ddag\dag}$ \\
		\bottomrule
		\multicolumn{10}{l}{$\S$: W (weight), N (neuron), H (attention head), E (embedding dimension), S (sublayer), L (layer), T (token)}  \\
		\multicolumn{10}{l}{\ding{168}: Bb (Bert-base), Rb (RoBERTa-base), \ding{169}: reported results in CoFi~\cite{cofi}, $\dag$: reported results in \kprune~\cite{kpruning}} \\
		\multicolumn{10}{l}{
		\fcr: FLOPs-based compression rate, $\ddag$: maximum speed up when accuracy drop < 3\%p} \\
		\multicolumn{10}{l}{
		\dif: (the accuracy (or F1 score) of a compressed model) - (the accuacy (or F1 score) of an uncompressed model)}
	\end{tabular}
	\vspace{-0.3cm}
\end{table}
\setlength{\tabcolsep}{6pt} 

\subsection{Pruning Granularity: Unstructured vs. Structured}
\label{subsec:p2}

Pruning granularity refers to the size of units that we use for pruning and we provide an illustration of pruning granularities regarding weights and tokens in Fig.~\ref{fig:3_granularity}.
$\mW^Q$, $\mW^K$, $\mW^V$, and $\mW^O\in\bR^{d\times d}$ are weights in an MHA sublayer. $\mU^I\in\bR^{d_f\times d}$ and $\mU^O\in\bR^{d\times d_f}$ are weights in an FFN sublayer.
$\vx_i$ and $\vx'_i$ are $i$th token embeddings before and after transformation through a transformer layer, respectively.
Colored dashed lines indicate the regions of weights included in each pruning granularity.
Pink-colored weights are related to the same embedding dimension.
The smallest granularity for pruning is an individual weight (W in Tables~\ref{tab:3_eo} and~\ref{tab:3_do}) for unstructured pruning~\cite{mvp, leap, obert,sparsegpt}.
On a larger scale, we use tokens (T), neurons (N), attention heads (H), embedding dimensions (E), sublayers (S), and layers (L) for structured pruning~\cite{dynabert,snip,ebert,bmp,cofi,sajjad23,ziplm,KKMHKG22,kcm,kpruning,pruner}.
Larger pruning granularities lead to speedup in inference, but, there is an accuracy degradation since it removes important weights along with unimportant weights.
The number of weights does not change after pruning tokens and it reduces only the number of FLOPs.

Unstructured pruning algorithms~\cite{mvp,leap,obert,cap} have demonstrated high accuracy in compressing language models achieving up to 97\% compression rate with a small accuracy degradation.
However, their drawback lies in the acceleration difficulty due to the resulting sparse pruning patterns.
Therefore, researchers have focused on extending the findings in unstructured pruning algorithms to structured pruning algorithms to achieve acceleration.
For example, BMP~\cite{bmp} extends the findings of MvP~\cite{mvp} considering that the magnitude of gradient of an objective function with respect to a weight is more important than the magnitude of the weight.
Similarly, ZipLM~\cite{ziplm} achieves a good performance on structured pruning of BERT by developing the findings in oBERT~\cite{obert} that the optimal error compensation strategy of OBS~\cite{obs} is useful for pruning BERT.

When extending unstructured pruning algorithms to structured pruning algorithms, it is crucial to determine the granularity for pruning, considering the characteristics of the model.
Michel et al.~\cite{sixteen} experimentally demonstrate that MHA sublayers are robust to the pruning of attention heads by showing that approximately 40\% of attention heads are prunable with negligible accuracy degradation.
Similarly, BMP~\cite{bmp} discovers the pattern that parameters in certain attention heads in MHA sublayers are entirely removed when we perform block-wise pruning with 16x16 weight blocks.
For FFN sublayers, the weight blocks within specific neurons in intermediate layers are entirely removed, while weight blocks in other neurons are preserved.
Based on these findings, subsequent structured pruning algorithms~\cite{dynabert,ebert,snip,cofi,cap,ziplm,KKMHKG22,kcm,kpruning,pruner} employ attention heads and intermediate neurons as the pruning granularities for MHA and FFN sublayers, respectively.
When we prune attention heads and neurons, we reformulate $M(\mX)$ and $F(\mX)$ in Equations~\eqref{eq:2_mha} and~\eqref{eq:2_ffn} into $M(X;\vzeta)$ and $F(X;\vxi)$ in Equation~\eqref{3:maskedmf} by imposing mask vectors $\vzeta \in \mathbb{R}^{H}$ and $\vxi \in \mathbb{R}^{N}$ that indicate the pruning status of $H$ attention heads and $N$ neurons, respectively; $\zeta_i=0$ represents that the $i$th attention head is pruned.
$\mW^O_i \in \bR^{d \times d_h}$ is the subset of $\mW^O$ corresponding to the $i$th attention head, i.e. from the $(d_h(i-1)+1)$th to the $(d_h i)$th columns of $\mW^O$.
$\mU^I_{i,:}\in\bR^{1\times d}$ and $\mU^O_{:,i}\in\bR^{d\times 1}$ are weights corresponding to the $i$th neuron in FFN sublayers.
We horizontally stack biases $\vb^O\in\bR^{d}$, $c_i^I\in\bR$, and $\vc^O\in\bR^{d}$ as $\mB^O = [\vb^O,\vb^O,...,\vb^O]\in\bR^{d \times s}$, $\mC_i^I=[c_i^I,c_i^I,...,c_i^I] \in\bR^{1 \times s}$, and $\mC^O=[\vc^O,\vc^O,...,\vc^O] \in\bR^{d \times s}$ to match the dimension in the equation where $s$ is a sequence length.
$c_i^I\in\bR$ is a bias corresponding to the $i$th neuron.
\begin{equation} \label{3:maskedmf}
M(X;\vzeta) = \sum_{i=1}^{H} \left(\zeta_i \mW^O_i \mA_i \right)+\mB^O
\; and \;
F(X;\vxi) = \sum_{i=1}^{N} \left(\xi_i \left( \mU^O_{:,i}
\sigma(\mU^I_{i,:} \mX +\mC^I_i) \right)\right)+\mC^O
\end{equation}
%
\begin{table}[t]
	\caption{Comparison of pruning algorithms for decoder-only Transformers on WikiText2 datasets.
		Lower perplexities represent better results.}
	\label{tab:3_do}
	\begin{tabular}{@{}llccrrrc@{}}
		\toprule
		\multicolumn{1}{c}{\multirow{2}{*}{\textbf{Method}}} & \multicolumn{1}{c}{\multirow{2}{*}{\textbf{PLM}}} & \multirow{2}{*}{\textbf{\begin{tabular}[c]{@{}c@{}}Pruning\\ Granularity$^\S$\end{tabular}}} & \multirow{2}{*}{\textbf{\begin{tabular}[c]{@{}c@{}}Pruning\\ Cost\end{tabular}}} & \multicolumn{1}{c}{\multirow{2}{*}{\textbf{\begin{tabular}[c]{@{}c@{}}Comp.\\ Rate\end{tabular}}}} & \multicolumn{2}{c}{\textbf{Perplexity}}     &
		\multicolumn{1}{c}{\multirow{2}{*}{\textbf{Speedup}}} \\
		\multicolumn{1}{c}{}   & \multicolumn{1}{c}{}   & & & \multicolumn{1}{c}{}    & \multicolumn{1}{c}{\textbf{PLM}} & \multicolumn{1}{c}{\textbf{Comp.$^\text{\comp}$}} &    \\
		\midrule
		SIMPLE~\cite{simple}   & GPT2-124M$^*$    & E, N, H     & High  & 48.5\%$^*$     & 14.49 & 17.14   & -     \\
		ZipLM~\cite{ziplm}     & GPT2-85M      & N, H, S     & High  & 44.4\%      & 24.10 & 30.30   & \multicolumn{1}{r}{1.50$\times$}      \\ \midrule
		Wanda~\cite{wanda}$^\natural$      & LLaMA-7B      & W     & Low   & 70.0\%      & 5.68  & 85.77   & -     \\
		SparseGPT~\cite{sparsegpt}$^\natural$    & LLaMA-7B      & W     & Low   & 70.0\%      & 5.68  & 26.30   & -     \\
		SparseGPT+OWL~\cite{owl}$^\natural$      & LLaMA-7B      & W     & Low   & 70.0\%      & 5.68  & 19.49   & -     \\
		\midrule
		SparseGPT~\cite{sparsegpt}   & OPT-175B      & 2:4      & Low   & 50.0\%      & 8.35  & 8.74    & \multicolumn{1}{r}{1.66$\times$$^\dag$}  \\
		LLM-Pruner~\cite{pruner}      & LLaMA-7B      & N, H     & Low   & 20.0\%      & 12.62 & 17.37   & \multicolumn{1}{r}{1.18$\times$}      \\
		\bottomrule
		\multicolumn{8}{l}{$\S$: W (weight), 2:4 (2:4 sparsity), N (neuron), H (attention head), S (sublayer)} \\
		\multicolumn{8}{l}{$\natural$: reported results in OWL~\cite{owl},
			*: includes token embeddings, \comp: a compressed model}       \\
		\multicolumn{8}{l}{$\dag$: geometric mean of speedups in individual affine transformations}
	\end{tabular}
	\vspace{-0.1cm}
\end{table}

As a result, the outputs of an MHA sublayer and an FFN sublayer are represented as the summation of outputs of independent attention heads and neurons.
Therefore, the outputs do not change when we remove attention heads or neurons whose mask variable is zero.

However, we cannot fully exploit GPU parallelism by pruning attention heads and neurons since the number of matrix multiplication between input and weight matrices is rarely reduced;
	pruning attention heads or neurons reduces the size of weight matrices, but it rarely removes the weight matrices entirely.
Therefore, we need to enlarge the pruning granularity to entire sublayers or even layers to maximize the inference speed of the compressed models.
Sajjad et al.~\cite{sajjad23} compares the accuracy of the compressed models with diverse layer pruning patterns, and finds that removing top layers is effective for encoder-only models.
CoFi~\cite{cofi} and ZipLM~\cite{ziplm}, which aim to prioritize acceleration performance, additionally prune the entire sublayers and achieve a good speedup.

Another approach involves the pruning of tokens.
These algorithms are designed based on the observation that the cosine similarity between tokens increases significantly as we move toward the top layers of BERT~\cite{powerbert}.
PoWER-BERT~\cite{powerbert} and LTP~\cite{ltp} achieve speedup more than two times with minimal accuracy degradation by using only a subset of tokens from the input sequence;
token pruning algorithms do not reduce the number of weights since the number of weights depends on the dimension of token embeddings not on the number of tokens.
While token pruning algorithms do not reduce the number of weights, their integration with other structured pruning algorithms employing different granularities such as CoFi~\cite{cofi} or ZipLM~\cite{ziplm} holds the potential for achieving remarkable speedup gains.

We use the same pruning granularity for decoder-only models as for encoder-only models, e.g.,  weight, attention heads, or neurons.
However, pruning decoder-only models is much more difficult than encoder-only models.
For instance, structured pruning algorithms~\cite{simple,ziplm} for GPT-2 show performance degradation at a low compression rate smaller than 50\% while those~\cite{bmp,cofi,ziplm} for BERT succeed in pruning up to 75\% of weights with minimal accuracy degradation.
Structured pruning approaches~\cite{pruner} for billion-scale LLMs show harsher accuracy degradation at a low compression rate of 20\% since the excessive size of LLMs limits the use of fine-tuning processes.
SparseGPT~\cite{sparsegpt} demonstrates relatively better performance using an unstructured pruning pattern and expands its algorithm to cover 2:4 pruning granularity which supports acceleration on customized hardwares; 2:4 pruning granularity represents a semi-structured pruning pattern that there are 2 zeros in 4 consecutive elements.
Improving the accuracy of structured pruning with large granularities for decoder-only models is a promising future work.

\subsection{Pruning Strategies: High-cost vs. Low-cost}
\label{subsec:p3}
\begin{wrapfigure}{r}{0.32\linewidth}
	\vspace{-0.33cm}
	\includegraphics[width=0.99\linewidth]{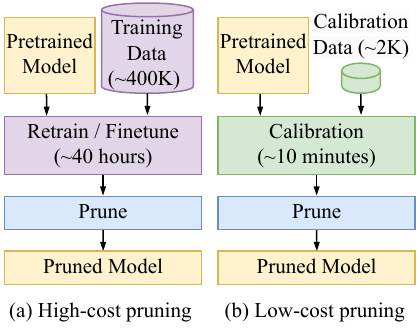}
	\caption{
		Comparison of high-cost and low-cost pruning algorithms.
		The size of datasets and the running times are based on the pruning algorithms~\cite{kpruning,dynabert} for BERT on the MNLI dataset.
	}
	\vspace{-0.5cm}
	\label{fig:3_cost}
\end{wrapfigure}

After selecting the granularity for pruning, we need to determine the strategy for pruning which includes methodologies of identifying unnecessary components and compensating for pruning errors induced by pruning of the identified components.
We categorize pruning algorithms into two groups of high-cost~\cite{mvp,leap,obert,dynabert,snip,ebert,bmp,cofi,cap,sajjad23,ziplm,powerbert,ltp,simple} and low-cost~\cite{KKMHKG22,kcm,kpruning,sparsegpt,wanda,owl,pruner} algorithms according to the cost of pruning.
We give an overview of high-cost and low-cost pruning algorithms in Fig.~\ref{fig:3_cost} where the size of datasets and the running time are based on the pruning algorithms~\cite{kpruning,dynabert} for BERT on the MNLI dataset.
High-cost pruning algorithms (a) generate an accurate compressed
 model through a computation-intensive retraining process on a large training dataset.
However, high-cost pruning algorithms are intractable requiring an excessive cost for pruning massive-scale LLMs, and low-cost algorithms (b) address the issue by substituting an expensive retraining process with a lightweight calibration process on a small calibration dataset.


We categorize high-cost pruning algorithms into three groups: regularized training-based algorithms~\cite{mvp,leap,bmp,flop,cofi}, OBS~\cite{obs}-based algorithms~\cite{obert,ziplm}, and others~\cite{dynabert,ebert,cap,snip}.
Regularized training-based algorithms introduce sparsity regularizations which regularize the magnitude of weights and gradually remove unnecessary components during a retraining process.
MvP~\cite{mvp} introduces a scoring matrix $\mS$ for each weight matrix $\mW$ and substitutes the weight matrix with $\mM \odot \mW$ where $\mM=(\mS>\tau)$ is a matrix of pruning masks that indicates whether the score of each weight exceeds the threshold $\tau$ or not.
$\odot$ is a Hadamard product.
MvP introduces a regularization term $R(\mS) = \lambda \sum_{i,j} \sigma_{s}\left(\mS_{i,j} \right)$ that penalizes the scores of weights to promote sparsification where $\sigma_{s}(\cdot)$ is a sigmoid function.
MvP manually controls the sparsity of the pruned model via tuning the coefficient $\lambda$.
LEAP~\cite{leap} addresses the inefficiency of manual selection of $\lambda$ by proposing an algorithm that automatically controls the sparsity level to match the preset sparsity.
BMP~\cite{bmp} extends MvP into structured pruning to get speedup in inference by sharing scores of the weights in the same group defined by the structural pruning granularity.

Another approach for regularized training is to impose $l_0$ regularization.
Let $\vw$ be a vector containing all of the weights in a PLM, then its $l_0$ norm $||\vw||_0$ is the number of non-zero weights which is same the as the size of the model.
Reducing the $l_0$ norm is equal to the model sparsification, but we can not directly impose $l_0$ regularization for training since it is not differentiable.
Louizos et al.~\cite{l0reg} addresses the problem by regularizing the expectation of the $l_0$ norm of weights multiplied by stochastic mask variables $\vz$, i.e. regularize $\mathbb{E}_{\vz} \left[ ||\vw \odot \vz||_0 \right]$ rather than $||\vw||_0$.
The modified regularization term is differentiable since the expectation of $l_0$ norm is computed as a closed form which is differentiable (see Louizos et al.~\cite{l0reg} for details).
FLOP~\cite{flop} utilizes the stochastic mask variables for structured pruning of language models and substitutes the $l_0$ regularization with an augmented Lagrangian to ensure the compressed model has a preset sparsity.
CoFi~\cite{cofi} achieves both high accuracy and fast inference speed via jointly pruning coarse and fine-grained granularities using an augmented Lagrangian in FLOP~\cite{flop}.

Given a weight vector $\vw$ and an input vector $\vx$, OBS~\cite{obs} algorithm finds an optimal weight-update $\delta\vw_{(q)}$ after pruning $q$th weight $w_q$ that minimizes the amount
$|| \vw \vx - \hat{\vw} \vx ||_2^2$ of a pruning error where $\hat{\vw}=\vw+\delta\vw_{(q)}$ is a vector of weights after pruning.
OBS is applied to language model compression by formulating the compression problem as a sublayer-wise output reconstruction problem after pruning~\cite{sparsegpt,ziplm}.
OBS is crucial for pruning PLMs since OBS-based algorithms demonstrate the state-of-the-art performance in both settings of high-cost~\cite{obert,ziplm} and low-cost pruning algorithms~\cite{sparsegpt}.
oBERT~\cite{obert} modifies the objective function of OBS into the cross-entropy and proposes an efficient algorithm to approximate the inverse of Hessian which is the computational bottleneck of OBS.
ZipLM~\cite{ziplm} extends the OBS algorithm to structured pruning by increasing the pruning granularity of OBS from a single weight to consecutive columns, and achieves remarkable acceleration up to 15x by finding runtime-aware configuration on the target hardware.

DynaBERT~\cite{dynabert} presumes a PLM as an overlap of inherent models with various configurations of attention heads, neurons, and sublayers.
DynaBERT succeeds in training all of the inherent models in a single training process and obtains pruned models by sampling the inherent models that satisfy the desired compression rate.
EBERT~\cite{ebert} trains sublayer-wise predictor networks that predict useful heads and neurons to refine sublayer inputs, and selects useful attention heads and neurons proper for the input of the sublayer.
CAP~\cite{cap} introduces a contrastive learning framework~\cite{contrastive} to pruning, which aims to make the intermediate features of pruned models mimic both pre-trained and fine-tuned model's intermediate features.
SNIP~\cite{snip} prunes attention heads and FFN sublayers whose maximum output is small, and proposes a spectral normalization that boosts the performance of pruned models.
The approaches in \cite{dynabert,ebert,cap,snip} provide unique insights for a PLM pruning problem, but they exhibit lower performance compared to regularized training-based~\cite{cofi} or OBS-based~\cite{ziplm} pruning algorithms.

Low-cost pruning algorithms~\cite{KKMHKG22,kcm,kpruning} significantly reduce the pruning cost by 1) separating the pruning process into mask search and error compensation processes, and 2) replacing the costly retraining-based error compensation process with an efficient weight-update process on a small calibration dataset.
Kwon et al.~\cite{KKMHKG22} is the first work that proposes a low-cost pruning framework for language models.
Kwon et al. efficiently estimates the saliency of attention heads and neurons via the empirical Fisher Information Matrix (FIM) that reflects the amount of influence on the objective function (cross-entropy) following the derivations in OBD~\cite{obd}.
Kwon et al. formulate an efficient sublayer-wise weight-tuning process that takes shorter than a minute using linear solvers.
KCM~\cite{kcm} addresses the inefficiencies of the method of Kwon et al. which requires expensive gradient computations for calculating FIM; KCM evaluates the importance of neurons using their representative powers without computing gradients.
However, Kwon et al. and KCM show significant accuracy degradations at high compression rates due to their one-shot pruning process without iteration.
\kprune~\cite{kpruning} demonstrates the importance of an efficient iterative pruning process by achieving remarkable  improvement up to 58\% higher accuracy than previous works without losing the efficiency of low-cost pruning processes.

Decoder-only models have not been extensively researched regarding low-cost pruning algorithms, especially for LLMs~\cite{llama,llama2,opt}.
SparseGPT~\cite{sparsegpt} succeeds in pruning massive-scale LLMs up to 175B on a single A100 GPU in three hours by maximizing the efficiency of the OBS~\cite{obs} algorithm.
We select SparseGPT as the representative pruning algorithm considering their contribution to LLM pruning and elaborate on details in Section~\ref{subsec:p4}.
Wanda~\cite{wanda} shows comparable results with SparseGPT in a low compression rate of 50\% using a significantly efficient unstructured pruning process without weight update, but it shows severe accuracy degradation in a high sparsity regime~\cite{owl}.
OWL~\cite{owl} improves the accuracy of both Wanda and SparseGPT by replacing their sublayer-wise uniform pruning rates with the non-uniform pruning rates reflecting the outlier ratio in each sublayer.
For structured pruning of LLMs, LLM-pruner~\cite{pruner} prunes attention heads and neurons using a Fisher Information matrix and uses LoRA~\cite{lora} for error compensation.
However, LLM-pruner faces a significant accuracy degradation even at a low compression rate of 20\%.
Therefore, structured pruning algorithms for LLMs have much room for improvement. 

\subsection{Representative Algorithm: SparseGPT}
\label{subsec:p4}
In this section, we elaborate on the details of SparseGPT~\cite{sparsegpt} which is the first algorithm that successfully prunes LLMs.
SparseGPT extends its unstructured pruning methodology to semi-structured pruning granularity, such as 2:4 sparsity, which supports acceleration in inference on NVIDIA A100 GPUs.
SparseGPT focuses on exploiting the optimal weight update in OBS~\cite{obs} to increase the accuracy of the pruned model, and reduce the number of computations of the Hessian inverse which is a computational bottleneck of the solution of OBS to maximize the efficiency.

Consider a binary pruning mask matrix $\mM \in \mathbb{R}^{d'\times d}$ for a weight matrix $\mW \in \mathbb{R}^{d'\times d}$ that projects $d$ dimensional input token embeddings $\mX \in \mathbb{R}^{d \times s}$ to the dimension of $d'$.
The authors formulate the error compensation problem for the weight $\mW$ as a least-square problem to find an optimal weight $\widehat{\mW}$ that minimizes the change in outputs as in Equation~\eqref{eq:3_s1}.
The authors divide the problem into $d'$ independent row-wise problems on the right-hand side.
\begin{equation}\label{eq:3_s1}
\arg\min_{\widehat{\mW}} \llVert \mW\mX - \left(\mM \odot \widehat{\mW})\mX \right) \rrVert_{F}^2
 = \arg\min_{\widehat{\mW}} \sum_{r=1}^{d'} \llVert \mW_{r,:}\mX -
\left(\mM_{r,:} \odot \widehat{\mW}_{r,:})\mX \right) \rrVert_{2}^2
\end{equation}
SparseGPT finds the optimal weight update $\delta\vw_{(r,m)}\in\mathbb{R}^{d}$ to compensate for the error induced by the pruning of the $m$th weight $\mW_{r,m}$ in the $r$th row, and the corresponding pruning error $\epsilon_{(r,m)}\in\mathbb{R}$ following OBS as in Equation~\eqref{eq:3_s2}.
In other words, $\widehat{\mW}^*_{r,:}=\mW_{r,:}+\delta\vw_{(r,m)}^T$ is the optimal solution of the subproblem regarding the $r$th row when the $m$th element is pruned, and
$\epsilon_{(r,m)}=\llVert \mW_{r,:}\mX - \widehat{\mW}^*_{r,:}\mX \rrVert_{2}^2$.
\begin{equation} \label{eq:3_s2}
\delta\vw_{(r,m)} = - \frac{\mW_{r,m}}{\left[\mH_{(r)}^{-1}\right]_{mm}} \left[\mH_{(r)}^{-1}\right]_{:,m}
\; and \;
\epsilon_{(r,m)} = \frac{1}{2}\frac{\mW^2_{r,m}}{\left[\mH_{(r)}^{-1}\right]_{mm}} \; where \; \mH_{(r)}=2\mX_{(r)} \mX_{(r)}^T
\end{equation}
$\mH_{(r)}$ is the Hessian of the objective function regarding the $r$th row, and $\mX_{(r)}$ is the subset of input features whose corresponding weights have not been pruned in the $r$th row.
Computing $\mH_{(r)}^{-1}$ is the major bottleneck in Equation~\eqref{eq:3_s2}; SparseGPT reduces the computational cost by adopting a column-wise pruning methodology.
SparseGPT iteratively performs column-wise pruning; 
they identify unnecessary weights, 
and sequentially perform an optimal weight update process to prune the identified weights in each column.
The column-wise pruning is efficient based on the following reasons.
First, since the Hessian depends on a pruning mask, not weights, different rows have the same Hessian inverse.
Therefore, they need to compute the Hessian inverse only once for each column when they prune in a column-wise manner.
Second, there is an efficient technique that computes the Hessian inverse for the subsequent column using the Hessian inverse of the previous column~\cite{UOPTQ}.
As a result, SparseGPT achieves remarkable efficiency that prunes 175B LLMs in 3 hours on a single A100 GPU.

The authors demonstrate the compatibility of SparseGPT with the state-of-the-art quantization algorithm~\cite{UOPTQ} and show that the combined algorithm shows better results than using only the quantization algorithm.
However, there are drawbacks of SparseGPT that come from sublayerwise pruning.
The pruning mask found by SparseGPT has uniform pruning rates for all sublayers missing sublayer-wise sensitivity.
OWL~\cite{owl} addresses the drawback by adjusting the pruning rates of sublayers based on their outlier ratios and achieves notable accuracy improvements.
The objective function of SparseGPT is a least-square in sublayer-wise outputs which is an indirect measure for the amount of errors in cross-entropy.
Therefore, the optimal solution in Equation~\eqref{eq:3_s2} might not be optimal for the direct objective function, and it would be improved when we directly reflect cross-entropy in the loss function. 

\section{Quantization}
\label{sec:quantization}
In this section, we introduce quantization algorithms that reduce the bit-width of parameters to compress large-scaled models while retaining the accuracy.
We first provide an overview in Section~\ref{subsec:Qoverview}, and describe various quantization schemes featured by quantization steps and levels in Section~\ref{subsec:Qlevel}, followed by quantization strategies in Section~\ref{subsec:Qcost}.
We single out OPTQ~\cite{UOPTQ} as the representative quantization algorithm for LLM compression, and provide detailed explanation of OPTQ and its improvement in Section~\ref{subsec:Qalg}.
\subsection{Overview}
\label{subsec:Qoverview}
Quantization is a core model compression technique that assigns continuous values distributed broadly to a small and discrete set.
Since Bhandare et al.~\cite{UBhandare2019} first made an attempt to quantize Transformer-based models, various methodologies~\cite{UZeroQuant2, UFineQuant, UOPTQ, URPTQ} have been proposed to quantize LLMs consisting of over a hundred billions of parameters.
Quantization minimizes memory footprint and inference latency of large-scaled models by reducing the number of bits required to represent the parameters.
Mapping continuous real values to their nearest integers is one of the most common approaches in neural network quantization.
This approach called round-to-nearest (RTN) takes less effort for assigning values to discrete space in virtue of its simplicity; however, naive RTN often suffers from severe performance degradation due to its limited representation capacity.
Other methodologies utilize non-uniform step sizes to quantize parameters for better representing the original ones.
Outputs of a quantizer, also referred to as \textit{quantization levels}, are also non-uniformly spaced under those methodologies in general.
Precision and datatype of quantization levels determine the compression rate and computational cost of quantized model, respectively.
It is straightforward that a lower bit-precision representation has a higher compression rate, and integer values benefits from accelerated operations.
We provide detailed explanation with regard to quantization step sizes and levels in Section~\ref{subsec:Qlevel}.

Meanwhile, cost of a quantization algorithm depends on quantization strategy to minimize accuracy degradation.
Quantization algorithms are computationally expensive when they entail the retraining process of an entire model.
Those high-cost algorithms, however, enable the resulting model to recover from the accuracy degradation caused by the low-precision representation.
\textit{Quantization-aware training (QAT)} is one of the representative quantization schemes that involve high-cost computation overhead to quantize models.
The prior quantization schemes for encoder-only Transformers including BERT~\cite{bert} are mostly based on QAT since the models have a relatively few parameters.
BERT quantization algorithms have focused on extreme quantization (under 4-bit), assisted by a fine-tuning process~\cite{UQ-BERT, UTernaryBERT, UBinaryBERT, UKDLSQ-BERT, UBiBERT, UMKQ-BERT, USensiMix, UBiT, UPreQuant}.
However, it is challenging to fully retrain the parameters of decoder-only Transformer, a mainstream architecture of LLMs.
Note that recently proposed LLMs~\cite{gpt3, opt, bloom} consist of hundreds of billions of parameters, as explained in Section~\ref{sec:prelim}.
This is why the low-cost quantization scheme including \textit{post-training quantization (PTQ)} is the most frequently utilized for LLM compression.
PTQ has drawbacks of severe accuracy degradation, but algorithms including~\cite{ULLMint8, USmoothQuant, UOPTQ, NUSqueezeLLM} succeed in preserving the performance under low-cost quantization scheme.
We explain more details of quantization strategies related to their cost in Section~\ref{subsec:Qcost}.

Tables~\ref{tab:QSummary_enc} and~\ref{tab:QSummary_dec} summarize quantization methodologies for encoder-only and decoder-only Transformers, respectively.
We categorize the algorithms based on the uniformity of quantization step size and the training cost, which we will describe in the following sections.
%
\begin{table}[t]
	\caption{Comparison of quantization algorithms for encoder-only Transformers on MNLI, QQP, and SQuAD$_{1.1}$ (SQD) benchmarks.}
	\label{tab:QSummary_enc}
		\begin{tabular}{@{}lcrccrrrrc@{}}
			\toprule
			\multicolumn{1}{c}{\multirow{2}{*}{\textbf{Method}}} & \multirow{2}{*}{\textbf{PLM$^\text{\ding{168}}$}} & \multirow{2}{*}{\textbf{Uniform$^\S$}} & \multirow{2}{*}{\textbf{Cost}} & \multirow{2}{*}{\textbf{\begin{tabular}[c]{@{}c@{}}Comp.\\ Rate\end{tabular}}} & \multicolumn{4}{c}{\textbf{Accuracy/F1 Difference$^\text{\dif}$ (\%p)}} &
			\multicolumn{1}{c}{\multirow{2}{*}{\textbf{Speedup}}} \\
			\multicolumn{1}{c}{} &  &  &  &  & \multicolumn{1}{c}{\textbf{MNLI}} & \multicolumn{1}{c}{\textbf{QQP}} & \multicolumn{1}{c}{\textbf{SQD}} & \multicolumn{1}{c}{\textbf{Avg.}} &  \\ \midrule
			Q8BERT~\cite{UQ8BERT} & Bb & UQ & High & 75.0\% & \multicolumn{1}{c}{-} & \multicolumn{1}{c}{-} & -0.72 & -0.72 & - \\
			TernaryBERT~\cite{UTernaryBERT} & Bb & UQ & High & 93.8\% & -1.20 & -0.80 & -1.30 & -1.10 & - \\
			QuantNoise~\cite{UQuantNoise} & Rb & UQ & High & 75.0\% & -1.20 & \multicolumn{1}{c}{-} & \multicolumn{1}{c}{-} & -1.20 & - \\
			BinaryBERT~\cite{UBinaryBERT} & DyB & UQ & High & 96.9\% & -1.00 & -0.20 & -2.50 & -1.23 & - \\
			I-BERT~\cite{UI-BERT} & Rl & UQ & High & 75.0\% & -0.30 & -0.20 & \multicolumn{1}{c}{-} & -0.25 & 4.00$\times$ \\
			SensiMix~\cite{USensiMix} & Bb & UQ & High & 87.5\% & \multicolumn{1}{c}{-} & -1.20 & \multicolumn{1}{c}{-} & -1.20 & 4.82$\times$ \\
			BiT~\cite{UBiT} & Bb & UQ & High & 96.9\% & -5.40 & -6.00 & -14.80 & -8.73 & - \\
			PreQuant~\cite{UPreQuant} & Rl & UQ & High & 87.5\% & -0.80 & -1.10 & \multicolumn{1}{c}{-} & -0.95 & - \\ \midrule
			K-means Quant.~\cite{NUK-meansQuant} & Bb & NUQ & High & 90.6\% & -0.60 & -0.40 & \multicolumn{1}{c}{-} & -0.50 & - \\ \midrule
			MREM~\cite{UMREM} & Bl & UQ & Low & 93.8\% & -3.10 & \multicolumn{1}{c}{-} & -3.70 & -3.40 & - \\
			Outlier Suppression~\cite{UOutlierSuppression} & Bb & UQ & Low & 81.3\% & -6.40 & -6.25 & -3.80 & -5.48 & - \\ \midrule
			GOBO~\cite{NUGOBO} & Bl & NUQ & Low & 89.8\% & -0.69 & \multicolumn{1}{c}{-} & -0.91 & -0.80 & - \\
			Mr.BiQ~\cite{NUMr.BiQ} & Bb & NUQ & Low & 93.6\% & -0.86 & \multicolumn{1}{c}{-} & -1.53 & -1.20 & - \\
			PowerQuant~\cite{NUPowerQuant} & Bb & NUQ & Low & 87.5\% & -1.52 & -0.16 & \multicolumn{1}{c}{-} & -0.84 & - \\ \bottomrule
			\multicolumn{10}{l}{\ding{168}: Bb/l (BERT-base/large), Rb/l (RoBERTa-base/large), DyB (DynaBERT)} \\
			\multicolumn{10}{l}{$\S$: UQ (Uniform quantization), NUQ (Non-uniform quantization)}\\
			\multicolumn{10}{l}{
			\dif: (the accuracy (or F1 score) of a compressed model) - (the accuacy (or F1 score) of an uncompressed model)} \\			
		\end{tabular}
\end{table}
%
\vspace{-0.3cm}
\subsection{Quantization Steps and Levels}
\label{subsec:Qlevel}
\begin{table}[t]
	\caption{Comparison of quantization algorithms for decoder-only Transformers on WikiText2 dataset. 
		Lower perplexities represent better results.
	}
	\label{tab:QSummary_dec}
	\begin{tabular}{@{}llrccrrc@{}}
		\toprule
		\multicolumn{1}{c}{\multirow{2}{*}{\textbf{Method}}} & \multicolumn{1}{c}{\multirow{2}{*}{\textbf{PLM}}} & \multirow{2}{*}{\textbf{Uniform}} & \multirow{2}{*}{\textbf{Cost}} & \multirow{2}{*}{\textbf{\begin{tabular}[c]{@{}c@{}}Comp.\\ Rate\end{tabular}}} & \multicolumn{2}{c}{\textbf{Perplexity}} &
		\multicolumn{1}{c}{\multirow{2}{*}{\textbf{Speedup}}} \\
		\multicolumn{1}{c}{} & \multicolumn{1}{c}{} &  &  &  & \multicolumn{1}{c}{\textbf{PLM}} & \multicolumn{1}{c}{\textbf{Comp.}$^\text{\comp}$} &  \\ \midrule
		Wu et al.~\cite{UWu2023} & GPT-2-medium & UQ & High & 87.5\% & 21.02 & 25.99 & \multicolumn{1}{r}{8.50$\times$} \\
		PEQA~\cite{UPEQA} & LLaMA-65B & UQ & High & 81.3\% & 3.53 & 4.27 & - \\
		LLM-QAT~\cite{ULLM-QAT} & LLaMA-30B & UQ & High & 87.5\% & 7.00 & 7.70 & - \\
		INT2.1~\cite{UINT2.1} & LLaMA-7B & UQ & High & 93.8\% & 5.08 & 8.74 & - \\ \midrule
		QLoRA~\cite{NUQLoRA} & LLaMA2-13B & NUQ & High & 87.5\% & 4.88 & 5.22$^\dag$ & - \\ \midrule
		ZeroQuant~\cite{UZeroQuant}$^\text{\ding{169}}$ & BLOOM-176B & UQ & Low & 81.3\% & 8.11 & 8.35 & - \\
		FineQuant~\cite{UFineQuant} & OPT-175B & UQ & Low & 87.5\% & 9.08 & 9.84 & \multicolumn{1}{c}{0.84$\times$} \\
		SmoothQuant~\cite{USmoothQuant} & LLaMA-65B & UQ & Low & 75.0\% & 6.17 & 6.20 & \multicolumn{1}{c}{1.36$\times$} \\
		OPTQ~\cite{UOPTQ}$^\natural$ & OPT-175B & UQ & Low & 81.3\% & 8.34 & 8.68 & \multicolumn{1}{c}{3.20$\times$} \\
		RPTQ~\cite{URPTQ} & OPT-175B & UQ & Low & 81.3\% & 8.34 & 10.03 & - \\
		AWQ~\cite{UAWQ} & LLaMA2-70B & UQ & Low & 81.3\% & 3.32 & 3.74 & \multicolumn{1}{c}{3.90$\times$} \\
		OWQ~\cite{UOWQ} & OPT-66B & UQ & Low & 80.6\% & 9.34 & 9.31 & - \\
		SpQR~\cite{USpQR} & LLaMA-65B & UQ & Low & 75.6\% & 3.53 & 3.68 & \multicolumn{1}{c}{1.29$\times$} \\ \midrule
		SqueezeLLM~\cite{NUSqueezeLLM} & LLaMA-13B & NUQ & Low & 78.0\% & 5.09 & 5.60 & \multicolumn{1}{c}{2.40$\times$} \\ \bottomrule
		\multicolumn{8}{l}{\ding{169}: reported results in ZeroQuant-V2~\cite{UZeroQuant2}, $\dag$: reported results in LoftQ~\cite{NULoftQ}, $\natural$: identical to GPTQ~\cite{UGPTQ}} \\
		\multicolumn{8}{l}{\comp: a compressed model} \\
	\end{tabular}
\end{table}
\subsubsection{Uniformity}
Quantization techniques are categorized into \textit{uniform quantization} and \textit{non-uniform quantization} according to the uniformity of quantization step sizes.
Uniform quantization generally maps a real-valued input to the nearest integer, which results in the output values to be uniformly spaced.
Non-uniform quantization, in contrast, leverages mapping functions that map input values to non-uniform space.
While uniform quantization is easy to be implemented, the reconstruction error of quantized model is minimized under non-uniform quantization scheme.
We provide more detailed explanation about uniform and non-uniform quantization in the following.
\paragraph{Uniform quantization}
The uniform quantizer $Q_U(\cdot)$ is defined as Equation~\eqref{eq:uniquant}, where $\lfloor\cdot\rfloor$ is a floor function, $s$ is a scaling factor, and $z$ is an integer offset.
Given an input matrix $\mW$, we
suppose $b$-bit quantization using the range of clipped inputs as $[\alpha,\beta]$ to calculate $s$ and $z$.
\begin{equation}
		Q_U(\mW)
	=\left\lfloor\frac{\mW}{s}+\frac{1}{2}\right\rfloor+z
	\text{,\quad where~}s=\frac{\beta-\alpha}{2^b-1}\text{~and~}
	z=\left\lfloor-\frac{\beta\cdot2^{b-1}+\alpha\left(2^{b-1}-1\right)}{\beta-\alpha}+\frac{1}{2}\right\rfloor\text{.}
	\label{eq:uniquant}
\end{equation}
\begin{wrapfigure}{r}{0.620\linewidth}
	\vspace{-0.57cm}
	\includegraphics[width=\linewidth]{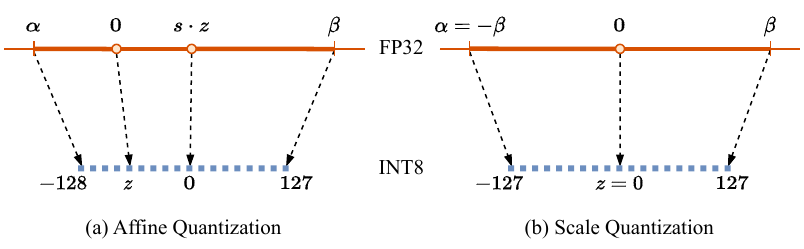}
	\vspace{-0.8cm}
	\caption{Comparison of 8-bit affine and scale quantization.}
	\vspace{-0.3cm}
	\label{fig:Qsymmetry}
\end{wrapfigure}
The clipping range $[\alpha,\beta]$ determines the scaling factor and the offset.
Choosing $\alpha$ and $\beta$, also referred to as \textit{calibration}, needs to be done adequately for decent representation of original real-valued inputs.
The basic and simple calibration is selecting minimum or maximum value of an input matrix.
However, the min/max approach is vulnerable to the input data with outliers, and results in poor representation of the original inputs.
One method to tackle the problem is using $k$-th percentile of inputs; the other is minimizing KL divergence between real-valued inputs and quantized outputs~\cite{Calibration}.

Uniform quantization methods are sub-categorized into affine quantization and scale quantization based on the symmetry of clipping range.
Fig.~\ref{fig:Qsymmetry} shows a brief comparison of those two different quantization schemes based on the calibration symmetry.
The absolute values of $\alpha$ and $\beta$ are different from each other in affine quantization; the integer offset $z$ is consequently non-zero, as illustrated in Fig.~\ref{fig:Qsymmetry} (a).
While affine quantization benefits from precise representation of original values, non-zero $z$ results in the computational overhead.
Scale quantization bypasses this problem by using symmetric clipping range.
Note that the range of quantized values accordingly needs to be symmetric (i.e., $[-2^{b-1}+1,2^{b-1}-1]$).
We redefine the quantizer for scale quantization as follows:
\begin{equation*}
	Q_U(\mW)
	=\left\lfloor\frac{\mW}{s}+\frac{1}{2}\right\rfloor,\quad\text{where}~s=\frac{\beta}{2^{b-1}}.
\end{equation*}
Affine quantization is recommended when the target real-valued inputs have the skewed distribution, while scale quantization is preferred if reducing the computational cost is the most important factor.
\begin{figure}
	\includegraphics[width=\textwidth]{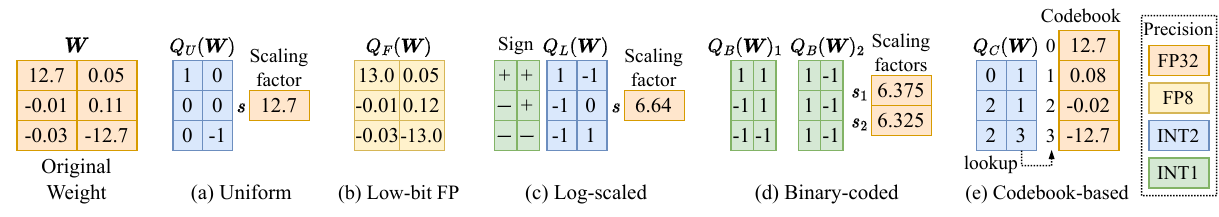}
	\caption{
		An example of different quantization schemes under 
		(a) naive round-to-nearest (RTN) uniform quantization,
		(b) non-uniform quantization using low-bit floating point,
		(c) non-uniform quantization using log-scaled values,
		(d) binary-coded non-uniform quantization, and
		(e) codebook-based non-uniform quantization.
	}
	\label{fig:QUniform}
\end{figure}
\paragraph{Non-uniform quantization}
Although the majority of proposed PLM compression schemes are founded on uniform quantization, non-uniform quantization~\cite{NUK-meansQuant, NUGOBO, NUMr.BiQ, NUPowerQuant, NUAlphaTuning, NUSqueezeLLM, NUNUPES} also have been spotlighted in behalf of its improved representation of the original parameters over uniform quantization~\cite{N2UQ}.
Uniformly quantizing NN models leads to significant accuracy degradation since most of parameters do not follow uniform distribution.
Assigning different size of quantization steps to each interval enables the quantized model to adequately represent the original parameters.
We categorize non-uniform quantization schemes into low-bit floating point, log-scaled, binary-coded, and codebook-based ones.
Methods including~\cite{NUFP8Quant, NUMicikevicius2022, NUZeroQuantFP} utilize low-bit floating points to represent outputs, as illustrated in Fig.~\ref{fig:QUniform} (b).
They leverage the quantizer $Q_F(\cdot)$ that maps the original value to its nearest low-precision floating point value.
Log-scaled non-uniform quantization algorithms map input values to logarithmic-distributed space~\cite{NUJi2021, NUPowerQuant, NUNUPES}.
The quantizer $Q_L(\cdot)$ of log-scaled quantization algorithms is defined as Equation~\eqref{eq:logquant}, where $Q_U(\cdot)$ is the uniform quantizer, $|\mW|$ denotes the matrix consisting of the absolute values of $\mW$, and the logarithmic base is $2$.
\begin{equation}
	Q_L(\mW) = Q_U( \log_2 { |\mW| } ).
	\label{eq:logquant}
\end{equation}
Note that $\mW \approx \text{Sign} (\mW) 2^{ s \cdot Q_L(\mW) }$ where $s$ is a scaling factor.
Following the process described above, the original weight matrix in the leftmost part of Fig.~\ref{fig:QUniform} is quantized as Fig.~\ref{fig:QUniform} (c).
Approaches leveraging binary-coding~\cite{NUMr.BiQ, NUAlphaTuning} decompose the original parameters into full-precision scaling factors and binary matrices.
As illustrated in Fig.~\ref{fig:QUniform} (d), all elements of a single matrix share $b$ scaling factors under $b$-bit binary-coded quantization,
where $b$ is 2 in the example.
Binary-coded quantization algorithms focus on finding the appropriate scaling factor $s_i$ and its corresponding binary matrix $Q_B(\mW)_i$ such that $\mW\approx\sum_{i=1}^{b}{s_i \cdot Q_B(\mW)_i}$.
For example, the element at the first row and column of the quantized matrix in Fig.~\ref{fig:QUniform} (d) is calculated as follows:
\begin{equation*}
	\left[Q_B(\mW)_1\right]_{11} \cdot s_1 + \left[Q_B(\mW)_2\right]_{11} \cdot s_2 = 6.375 + 6.325 = 12.7 = \mW_{11}.
\end{equation*}
The other technique, named codebook-based quantization~\cite{NUK-meansQuant, UQuantNoise, NUGOBO, NUSqueezeLLM}, utilizes look-up tables and indices.
These methodologies are based on $k$-means clustering; the key point is to find $2^b$ centroids for $b$-bit quantization.
The quantizer $Q_C(\cdot)$ of codebook-based quantization maps the original inputs to indices pointing to their corresponding centroids.
Therefore,
the original matrix is decomposed to the codebook consisting of full-precision centroids and the matrix with low-bit indices, as in Fig.~\ref{fig:QUniform} (e).
%
%
%
\subsubsection{Precision and Type}
Precision of the resulting outputs is one of the most important features of quantizer, as well as datatype of outputs.
Output precision determines the compression rate; for example, the compressed model is $4\times$ smaller than the original one if the model with 32-bit floating points is quantized using an 8-bit quantization algorithm.
It is common to quantize FP32 models with INT8 precision, while recently proposed algorithms focus more on \textit{extreme quantization}.
Extreme quantization refers to techniques that reduce the bit precision under 4-bit~\cite{UTernaryBERT, UQuantGPT, ULLM-QAT, UINT2.1, UPreQuant, UOutlierSuppression, URPTQ, UAWQ, UQuIP, UMREM, UWu2023}.
However, quantizing the model with extremely low-bit precision representation is vulnerable to accuracy degradation.
One way to tackle the problem is using \textit{mixed-precision quantization} algorithms.
In Transformer-encoder model quantization,~\cite{USensiMix} leverages different bit-precision based on the sensitivity analysis.
Similar approaches including~\cite{UOWQ, USpQR} are applied to quantizing Transformer-decoder models.

Meanwhile, output datatype is related to the computation latency of the resulting model.
Computing hardwares, such as CPU or GPU, generally support acceleration of integer arithmetic.
In contrast, it is not common for hardwares to accelerate floating-point arithmetic operations.
Quantization algorithms are hard to benefit from the acceleration when they entail floating-point operations.
\textit{Dequantization} is the main reason why the quantized model is unable to be supported by integer arithmetic acceleration.
Dequantizing the discrete values is the exact opposite of quantization; that is, dequantization projects quantized outputs back to continuous real values.
%
Approaches including~\cite{UQ8BERT, UI-BERT} devise novel LM quantization schemes that do not entail a dequantization process, so that the resulting model can perform faster inference.
Some algorithms utilize different type of output;~\cite{UPEQA} adapts BFloat16 to store embedding and activation map, and~\cite{NUQLoRA} proposes a novel data type NormalFloat4. 
$\;$\\$\;$\\
\vspace{-1cm}
\subsection{Quantization Strategies: High-cost vs. Low-cost}
\label{subsec:Qcost}
Quantization cost is mainly decided based on whether the algorithms entail retraining or fine-tuning after quantization.
High-cost methods require entire or partial training of quantized models to recover from the accuracy degradation.
In contrast,
low-cost methods do not require the training of quantized models; thus, it is cheaper and faster to utilize low-cost algorithms than high-cost ones.
However, the severe accuracy degradation is the most critical drawback of low-cost algorithms.
We provide more detailed explanation of two main quantization strategies based on the training cost in the following.
\subsubsection{High-cost}
Original full precision models pass through the entire retraining or partial fine-tuning process under high-cost quantization schemes.
A general approach of language model compression,
called \textit{quantization-aware training (QAT)},
1) applies the quantization to pretrained Transformer-based models,
and 2) retrain or fine-tune the model to recover from the performance deterioration after quantization.
However, it is impossible to compute the gradients during back propagation due to the existence of non-differentiable step functions.
One of the most common methods
\begin{wrapfigure}{r}{0.370\linewidth}
	\includegraphics[width=\linewidth]{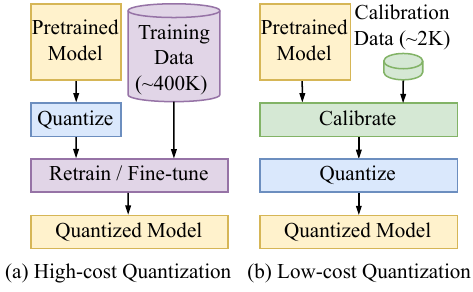}
	\vspace{-0.8cm}
	\caption{Comparison of high-cost and low-cost quantization.}
	\label{fig:QCost}
\end{wrapfigure}
 to tackle this problem is utilizing straight-through estimator (STE), as described in Section~\ref{sec:prelim}.
Although the functions for forward and backward pass are different from each other under STE, the algorithms still works well in practice.
Since high-cost algorithms require the training or fine-tuning dataset as illustrated in Fig.~\ref{fig:QCost} (a), it is difficult to be applied when those data are inaccessible.

High-cost quantization algorithms for BERT compression~\cite{UQ8BERT, UQuantNoise, UI-BERT, USensiMix} retrain the quantized models utilizing task-specific fine-tuning data.
Since BERT has plenty of redundancies~\cite{BERTRedundancy} and a relatively small number of parameters, most of BERT quantization algorithms reduce the bit-length under 2-bit~\cite{UQ-BERT, UTernaryBERT, UBinaryBERT, UBiT}, and retrain the quantized models to compensate for the quantization errors.
However, retraining LLMs is highly time- and memory-consuming due to their prohibitive number of parameters.
Recent high-cost algorithms propose a lightweight QAT to minimize the cost of model retraining, based on the fact that it is sufficient to fine-tune only the fraction of parameters for preserving the accuracy~\cite{UINT2.1}.
Lightweight QAT updates only some portion of parameters while keeping the other fixed.
Scaling factors~\cite{UPEQA}, parameters with high sensitivity~\cite{ULLM-QAT}, low-rank weight matrices~\cite{UINT2.1, NUQLoRA}, and outliers~\cite{UPreQuant} are the targets of lightweight QAT.

Unlike high-cost quantization algorithms,
it is also possible to minimize quantization cost by quantizing parameters without any retraining processes.
We provide the detailed explanation of those low-cost quantization algorithms in the following.

\subsubsection{Low-cost}

Low-cost quantization algorithms do not require time- and memory-intensive model retraining or fine-tuning.
Low-cost quantization is adequate to quantize models with a gigantic number of parameters; it consequently has become the mainstream approach to quantize LLMs, as shown in Table~\ref{tab:QSummary_dec}.
Low-cost quantization algorithms mostly follow \textit{post-training quantization (PTQ)}, a training-free quantization scheme.
PTQ quantizes pretrained Transformer-based models without retraining or fine-tuning the model after quantization.
As illustrated in Fig.~\ref{fig:QCost} (b), low-cost algorithms directly quantize the pretrained model; a tiny subset of the training data is needed only for calibration.
Low-cost quantization algorithms are sub-categorized into \textit{dynamic quantization} and \textit{static quantization}. 

Dynamic quantization~\cite{UZeroQuant, NUPowerQuant} determines the clipping range of activations to calculate scaling factors and integer offsets during inference.
%
Dynamic quantization algorithms generally show better performance than static algorithms since they utilize actual input data to compute scaling factors and integer offsets.
However, dynamic quantization algorithms result in slower inference speed of the resulting model, due to the extra computational overhead during inference.
Static quantization~\cite{URPTQ, USmoothQuant}, on the other hand, precomputes scaling factors and integer offsets of activations before inference.
Static quantization algorithms require a small subset of training data;
those partial data called \textit{calibration data} decide the clipping range of activation.
Although static methods for activation quantization are lightweight, they are not applied for most of LLM compression due to severe performance degradation.
The degradation in performance results from various range of actual input data, caused by diverse lengths of sequence and embedding values.

Activation quantization mainly aims to minimize the computational overhead of model inference; however, memory overhead dominates the entire inference time of LLMs, rather than the computational overhead~\cite{NUSqueezeLLM}.
In consequence, recent LLM quantization algorithms mainly focus on minimizing memory overhead of LLMs.
Weight-only quantization, which is regarded as a static algorithm, is one representative approach to effectively reduce the memory usage of LLMs.
Weight-only quantization algorithms~\cite{NUGOBO, ULLMint8, UOPTQ, UAWQ, UOWQ, USpQR, NUSqueezeLLM, UQuIP, NUNUPES} quantize only the weights of given models while keeping activations as full-precision values.
It is obvious that weight-only quantization algorithms involve dequantization processes during the inference; the quantized models therefore cannot be supported by integer arithmetic acceleration.
Nevertheless, weight-only quantization optimizes inference speed by minimizing the latency associated with loading models into memory.
Meanwhile, weight-only quantization algorithms normally do not require the calibration data; however, some weight-only quantization algorithms~\cite{UOPTQ, UAWQ} need a small number of calibration data to quantize weights. 
\subsection{Representative Algorithm: OPTQ}
\label{subsec:Qalg}

OPTQ~\cite{UOPTQ} resolves the accuracy degradation caused by naive round-to-nearest (RTN), and quantizes large-scaled LLMs~\cite{opt,bloom} to under 3 bits with negligible performance degradation.
Succeeding works~\cite{URPTQ, UOWQ, UQuIP, UQuantEase, UNormTweaking} have been developed based on OPTQ to improve the performance.
As the basic mechanism of OPTQ is identical to that of SparseGPT~\cite{sparsegpt} explained in Section~\ref{subsec:p4}, we focus more on how OPTQ is improved over time.

\begin{figure}[t]
	\includegraphics[width=\textwidth]{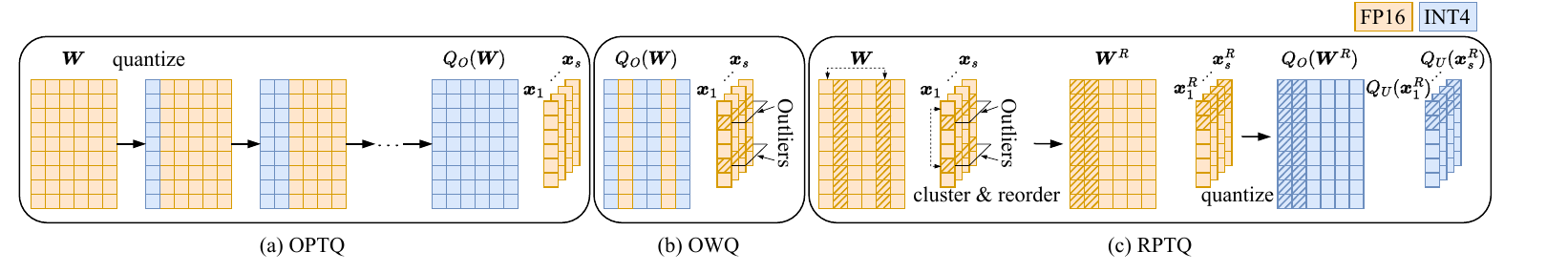}
	\caption{
		Illustrations of quantizing FP16 weights $\mW$ and token embeddings $\{\vx_i\}^s_{i=1}$  into INT4 values with OPTQ~\cite{UOPTQ} and its variants~\cite{URPTQ,UOWQ}, where $s$ is the sequence length.
		OWQ and RPTQ preserve the accuracy by carefully managing outliers represented as the dashed elements in $\{\vx_i\}^s_{i=1}$; dashed elements in $\mW$ represent outlier-related weights.
		The superscript $R$ denotes that a matrix or vector is reordered.
		$Q_O(\cdot)$ and $Q_U(\cdot)$ are the quantizers of OPTQ and round-to-nearest (RTN), respectively.
	}
	\label{fig:optq}
\end{figure}

The major difference between SparseGPT and OPTQ lies in the resulting values; a target weight $\mW_{r,m}$ becomes $0$ in SparseGPT while it becomes the corresponding quantized value in OPTQ.
The authors find the optimal weight update $\delta\vw_{(r,m)}$ to compensate for the error $\epsilon_{(r,m)}$ induced by quantizing the $m$th weight $\mW_{r,m}$ in the $r$th row as in Equation~\eqref{eq:obq}; the updated $r$th row $\widehat{\mW}_{r,:}$ is equal to $\mW_{r,:}+\delta\vw_{(r,m)}^T$.
\begin{equation}
	\delta\vw_{(r,m)} = - \frac{\mW_{r,m}-Q_U(\mW_{r,m})}{\left[\mH_{(r)}^{-1}\right]_{mm}} \left[\mH_{(r)}^{-1}\right]_{:,m}
	\; and \;
	\epsilon_{(r,m)} = \frac{1}{2}\frac{\left(Q_U(\mW_{r,m})-\mW_{r,m}\right)^2}{\left[\mH_{(r)}^{-1}\right]_{mm}} \; where \; \mH_{(r)}=2\mX_{(r)} \mX_{(r)}^T.
	\label{eq:obq}
\end{equation}
Similar to SparseGPT, OPTQ performs column-wise quantization 
to minimize the computations of $\mH_{(r)}^{-1}$.
Since quantization does not require mask searching, OPTQ directly quantizes the 
columns followed by error compensation in a column-wise manner.
We briefly depict the quantization process of OPTQ in Fig.~\ref{fig:optq} (a). 
Note that all the algorithms improving OPTQ quantize weight matrices following this process.

Among various improvements over OPTQ, we introduce two algorithms: OWQ~\cite{UOWQ} and RPTQ~\cite{URPTQ}.
Recent studies for LLM quantization~\cite{ULLMint8, USpQR, UAWQ} have shown that preserving activation outliers, which are significantly larger activation values than others, enhances the performance of the quantized models; quantizing activation outliers is the main cause of the accuracy degradation in the quantized LLMs.
Based on OPTQ algorithm, both OWQ~\cite{UOWQ} and RPTQ~\cite{URPTQ} take the outliers into account, effectively reducing the accuracy degradation.
OWQ is a weight-only quantization algorithm that enhances the performance by keeping the weights associated with the activation outliers in full precision, as depicted in Fig.~\ref{fig:optq} (b).
OWQ defines the sensitivity of each weight based on the activations; after finding the sensitive weights, OWQ quantizes the remaining insensitive weights with OPTQ algorithm.
Unlike OPTQ which focuses on quantizing only the weights of a given model, RPTQ applies activation quantization on top of OPTQ, as illustrated in Fig.~\ref{fig:optq} (c).
Note that RPTQ quantizes weight matrices with OPTQ algorithm, while utilizing RTN for quantizing activations.
RPTQ leverages multiple scaling factors and integer offsets in order to represent both outliers and non-outliers more accurately.
RPTQ groups the rows of activation matrix $\mX=[\boldsymbol{x}_1;\boldsymbol{x}_2;\cdots;\boldsymbol{x}_s]\in\bR^{d\times s}$ with similar value ranges, and individually quantizes each group with its optimal scaling factor and integer offset.
To leverage hardware-efficient block-wise quantization,
RPTQ reorders the activation matrices so that the consecutive rows within a block share the same scaling factor and integer offset;
the weight matrices are reorganized accordingly.

The critical limitation of OPTQ and its variants is that they do not reflect the dependency between sublayers when they update weights.
Furthermore, RPTQ suffers from an accuracy degradation problem when they quantize activation below 8 bits and OWQ suffers from slow inference speed because of the mixed precision weights.
We need to address those issues to expand the practical usage of quantized language models. 

\section{Other Algorithms}
\label{sec:others}

In this section, we describe other compression methods for LLMs, including knowledge distillation (KD), low-rank approximation (LRA), parameter sharing, and efficient architecture design.
%
%
We introduce a representative algorithm, LoRA~\cite{lora}, which fine-tunes a model by updating additional parameters in low-rank adaptors, rather than updating the original parameters.
A brief description of each technique is as follows:

\begin{itemize*}
	\item \textbf{Knowledge distillation} is a training technique that transfers knowledge from a pre-trained teacher model to a smaller student model, allowing the student model to mimic the teacher model.
	\item \textbf{Low-rank approximation} is a model compression technique that reduces the size of model by decomposing a high-dimensional matrix or tensor with lower-dimensional ones.
	\item \textbf{Parameter sharing} compresses a pre-trained model by utilizing identical weights in different parts of the model.
	\item \textbf{Efficient architecture design} involves the method of designing a Transformer layer with an efficient structure or automatically searching a model architecture from a pre-trained model that meets constraints.
\end{itemize*}

The rest of the section is organized as follows.
We describe KD, LRA, parameter sharing, and efficient architecture design in Sections~\ref{subsec:o2},~\ref{subsec:o3},~\ref{subsec:o4}, and~\ref{subsec:o5}, respectively.
Finally, we provide detailed explanations of LoRA in Section~\ref{subsec:o6}.

\subsection{Knowledge Distillation}
\label{subsec:o2}
Knowledge distillation (KD) is a technique that transfers useful knowledge from a large teacher model $\mathcal{T}$ to a small student model $\mathcal{S}$ to improve the accuracy of $\mathcal{S}$.
This process encourages the student model to generate outputs that are similar to that of the teacher model to obtain the teacher model's generalized knowledge.
KD matches the outputs of classifiers, embedding layers, and sublayers in $\mathcal{T}$ and $\mathcal{S}$ to distill diverse types of knowledge within different distillation sources.
We explain diverse types of distillation sources used for KD in Section~\ref{subsubsec:kd1}.

When we apply KD for language models, we need to consider how to match sublayers in $\mathcal{T}$ and $\mathcal{S}$ since they have different numbers of sublayers.
We categorize the sublayer matching strategies into three groups: 1:1, many:1, and many:many strategies.
We explain more details about each of the sublayer matching strategies in Section~\ref{subsubsec:kd2}.

We summarize the performance of diverse KD algorithms for encoder-only Transformers in Table~\ref{tab:5_kd}.
We denote the distillation sources and the sublayer-matching strategies of each KD algorithm in columns 2 and 3, respectively.
We report the amounts of accuracy/ F1 score degradation and acceleration after compression on MNLI, QQP, and SQuAD$_{1.1}$ benchmarks in columns from 5 to 9 where their compression rate is reported in the 4th column.
In this section, we focus on KD algorithms for encoder-only Transformers since there are lack of studies regarding KD algorithms for decoder-only Transformers.

\begin{table}[]
	\caption{
		Comparison of the performance of knowledge distillation (KD) algorithms for encoder-only Transformers on MNLI, QQP, and SQuAD$_{1.1}$ (SQD) benchmarks.
		We summarize the experimental results of KD algorithms distilling knowledge from a pretrained BERT-base to a compressed one with fewer sublayers; a 50\% compression rate means the student model contains the half of the number of sublayers of the teacher model (BERT-base)
	}
	\label{tab:5_kd}
		\begin{tabular}{lcccrrrrr}
			\toprule
			\multicolumn{1}{c}{\multirow{2}{*}{\textbf{Method}}} & \multirow{2}{*}{\textbf{\begin{tabular}[c]{@{}c@{}}Distillation\\ Source$^{\S}$\end{tabular}}} & \multirow{2}{*}{\textbf{\begin{tabular}[c]{@{}c@{}}Layer\\Matching \end{tabular}}} & \multirow{2}{*}{\textbf{\begin{tabular}[c]{@{}c@{}}Comp.\\ Rate\end{tabular}}} & \multicolumn{4}{c}{\textbf{Accuracy/F1 Difference$^\text{\dif}$ (\%p)}}   & \multirow{2}{*}{\begin{tabular}[c]{@{}c@{}}\textbf{Speedup}\end{tabular}} \\
			\multicolumn{1}{c}{} & & & & \textbf{MNLI} & \textbf{QQP} & \textbf{SQD$^*$} & \textbf{Avg.} & \\ \midrule
			DistilBERT~\cite{distilbert} & L & None & 50.00\% & -4.50 & -1.10 & -2.70 & -2.77 & - \\ \midrule
			MiniLM~\cite{minilm} & L, A & 1:1 & 50.00\% & -0.50 & -0.30 & -0.40$^\dagger$ & -0.40 & - \\
			MiniLMv2~\cite{minilm2} & L, A & 1:1 & 50.00\% & -0.30 & 0.20 & -0.50$^\dagger$ & -0.20 & 2.00$\times$ \\
			TinyBERT~\cite{tinybert}$^\ddag$ & L, H, A, T & 1:1 & 50.00\% & 0.70 & 0.50 & - & 0.60 & 2.00$\times$ \\
			TinyBERT~\cite{tinybert}$^\ddag$ & L, H, A, T & 1:1 & 66.70\% & -1.40 & 0.20 & - & -0.60 & 9.40$\times$ \\
			PKD~\cite{pkd} & L, H & 1:1 & 50.00\% & -2.20 & -0.10 & - & -1.15 & - \\
			Aguilar et al.~\cite{aguilar20} & L, H, A & 1:1 & 50.00\% & - & -0.07 & - & -0.07 & - \\ \midrule
			CKD~\cite{ckd}$^\text{\ding{169}}$ & L, H & many:1 & 50.00\% & -1.48 & -0.32 & - & -0.90 & -\\
			ALP-KD~\cite{alpkd} & L, H & many:1 & 50.00\% & -1.53 & -0.23 & - & -0.88 & - \\
			\midrule
			BERT-EMD~\cite{bertemd} & L, H, A, T & many:many & 50.00\% & 0.30 & 0.40 & - & 0.35 & 1.90$\times$ \\
			RAIL-KD~\cite{railkd} & L, H & many:many & 50.00\% & -1.20 & 0.40 & - & -0.40 & - \\ \bottomrule
			\multicolumn{9}{l}{$\S$: L (logits), H (hidden states), A (attention maps), T (token embeddings), $\ddag$: utilizes additional training data}\\
			\multicolumn{9}{l}{* $\text{SQuAD}_{1.1}$,
				$\dagger$: experimental results of RoBERTa-base on SQuAD$_{2.0}$,
				\ding{169}: reported results in ALP-KD~\cite{alpkd}} \\
			\multicolumn{9}{l}{
				\dif: (the accuracy (or F1 score) of a compressed model) - (the accuracy (or F1 score) of an uncompressed model)}
	\end{tabular}
\end{table}

\subsubsection{Distillation Source}
\label{subsubsec:kd1}
A distillation source is a set of information in $\cT$ and $\cS$ that are used for distilling knowledge.
KD uses logits (L), hidden states (H), attention maps (A), and token embeddings (T) as distillation sources.
Logits contain task-specific knowledge which directly affects the performance of language models, and all KD algorithms~\cite{distilbert,minillm,minilm2,tinybert,pkd,aguilar20,alpkd,railkd} in Table~\ref{tab:5_kd} include logits in their list of distillation sources.
KD distills inherent knowledge in logits by reducing the KL divergence between logits $\vz^\cT$ and $\vz^\cS$ as in Equation~\eqref{eq:logitkd}.
Superscripts $\cT$ and $\cS$ represent that the item is related to the teacher model and the student model, respectively.
$\text{D}_\text{KL}(\cdot)$ is a KL divergence function and $s_t(\cdot)$ is a softmax function with temperature $t$.
A softmax function with a higher temperature produces a softer probability distribution than that with a lower temperature.
\begin{equation}
\label{eq:logitkd}
\mathcal{L}_{\text{pred}} = \text{D}_{KL} (s_t(z^{\cT}) || s_t(z^{\cS}))
\end{equation}
KD matches the information of $\cT$ and $\cS$ in sublayers to distill generalized knowledge in sublayers.
Matching information in sublayers is vital since it gradually aligns the outputs of $\cS$ and $\cT$ from the lowest sublayer, finally aiding in the synchronization of the predictions between $\cS$ and $\cT$.
There are two distillation sources in Transformer sublayers: hidden states and attention maps.
PKD~\cite{pkd} distills the knowledge of hidden states by reducing the gap between hidden states $\mH^{\cS}_n$ and $\mH^{\cT}_m$ as in Equation~\eqref{eq:hidkd}.
$n$ and $m$ are indices of matched sublayers in $\cS$ and $\cT$, respectively.
We explain how we match those sublayers in Section~\ref{subsubsec:kd2}.
PKD introduces a weight $\mW_h$ to match the dimensions of $\mH_n^\cS$ and $\mH_m^\cT$.
\begin{equation}
	\label{eq:hidkd}
	\mathcal{L}_{\text{hid}}= \text{MSE}(\boldsymbol{H}_m^{\cT}, \mW_h \boldsymbol{H}_n^{\cS})
\end{equation}
A self-attention mechanism is the core of Transformer-based language models and an attention map contains abundant knowledge of how to capture contextual information.
The attention map $P_{i}$ of the $i$th attention head is computed as $S(K_i^T Q_i / d_h)$ in Equation~\eqref{eq:2_mha}.
The $j$th column $(P_{i})_{:,j}$ indicates how the $j$th token is affected by each token.
Aguilar et al.~\cite{aguilar20} reduces the gap between attention maps of $\cT$ and $\cS$ to transfer the knowledge within the attention map as in Equation~\eqref{eq:attkd}.
In this equation, $m$ and $n$ represent the indices of the sublayers of $\cT$ and $\cS$, respectively.
\begin{equation}
	\label{eq:attkd}
	\mathcal{L}_{\text{att}} =
	\sum_{i=1}^{H}\sum_{j=1}^{s}\text{D}_{KL} \left((\mP_{m,i}^{\cT})_{:,j}|| (\mP_{n,i}^{\cS})_{:,j} \right)
\end{equation}
TinyBERT~\cite{tinybert} distills the knowledge in token embeddings by reducing the gap between embedding matrices $\mE^{\cS}$ and $\mE^{\cT}$.
TinyBERT introduces a weight $\mW_e$ to match the dimensions of $\mE^\cS$ and $\mE^\cT$.
\begin{equation}
	\label{eq:embkd}
	\mathcal{L}_{\text{emb}}= \text{MSE}(\mE^{\cT}, \mW_e\mE^{\cS})
\end{equation}
%
\subsubsection{Layer Matching}
\label{subsubsec:kd2}

Layer matching strategies represent how to match sublayers of $\cT$ and $\cS$ to distill knowledge even though they have different numbers of sublayers.
We categorize sublayer matching strategies into 1:1, many:1, and many:many based on the number of sublayers used for knowledge distillation. 

1:1 sublayer matching strategy distills the knowledge from individual sublayers of $\cT$ to a corresponding sublayer in $\cS$.
Skip and Last mapping strategies are widely used to match the sublayers of $\cT$ and $\cS$~\cite{tinybert, pkd, aguilar20, lrcbert, mtbert, tutorkd, ted, jha2023}.
The skip mapping strategy presumes that the useful knowledge of $\cT$ is widely spread across its sublayers; thus, it matches each sublayer of $\cS$ with a sublayer of $\cT$ at uniform intervals.
In contrast, the Last mapping strategy presumes that the essential knowledge of $\cT$ is concentrated in the latter parts of $\cT$, and accordingly they sequentially match the uppermost sublayers of $\cT$ to those of $\cS$.
However, 1:1 sublayer matching strategies have the drawback that they lose knowledge within the unselected sublayers in $\cT$ since the number of sublayers of $\cT$ is larger than that of $\cS$ in most cases.

Many:1 sublayer matching strategy allows a sublayer of $\mathcal{S}$ to receive knowledge from multiple sublayers of $\mathcal{T}$.
CKD~\cite{ckd} groups the sublayers of $\mathcal{T}$ into multiple buckets and distills the knowledge within a bucket to a corresponding sublayer in $\cS$.
CKD resolves the knowledge loss problem of 1:1 matching strategies by letting all sublayers of $\cT$ participate in knowledge distillation.
However, they employ inaccurate heuristics for defining buckets and matching them to sublayers in $\cS$~\cite{alpkd}.
ALP-KD~\cite{alpkd} and Universal-KD~\cite{universialkd} employ an attention mechanism to learn the amount of knowledge that each sublayer in $\cS$ receives from each sublayer in $\cT$.
They do not suffer from inaccurate heuristic mappings of sublayers in $\cS$ and $\cT$ since they automatically learn an accurate mapping using the attention mechanism.
However, they require a long training time since they distill knowledge between all pairs of sublayers in $\cS$ and $\cT$.

Many:many sublayer matching strategy distills the knowledge from multiple sublayers of $\cT$ to those of $\cS$.
In many:many matching strategy, it is important to determine the matching pattern between sublayers of $\cT$ and those of $\cS$ since the matching pattern effectively enhances the performance of $\cS$.
BERT-EMD~\cite{bertemd} finds a matching pattern between sublayers of $\cT$ and $\cS$ by employing earth mover's distance (EMD).
Although the resulting model of BERT-EMD shows improved performance, BERT-EMD is time-intensive due to the computational overhead of searching the pattern.
RAIL-KD~\cite{railkd} adopts a lightweight approach compared to BERT-EMD; it 1) randomly selects sublayers of $\cT$ at each epoch, and 2) matches the selected sublayers with those of $\cS$.
RAIL-KD sets the number of the selected sublayers of $\cT$ the same as that of $\cS$.
RAIL-KD shows that the randomized selection enables all sublayers of $\cT$ to provide their knowledge to those of $\cS$, minimizing the computational overhead of finding the optimal pattern for sublayer matching.

\subsection{Low-Rank Approximation}
\label{subsec:o3}
Low-rank approximation (LRA) is a technique that decomposes a high-dimensional matrix (tensor) into a multiplication of lower-dimensional matrices (tensors) that approximates the original one.
LRA is effective for language models since parameters in language models have low-rank characteristics~\cite{exploring, collabo, linformer, drone}; the parameters are located in the low-dimensional subspace in the high-dimensional space.
We summarize the performance of LRA algorithms for encoder-only Transformers in Table~\ref{tab:lra} and divide them into two groups based on their target matrices for matrix (tensor) decomposition.

\begin{table}[t]
	\caption{Comparison of low-rank approximation for encoder-only Transformers. We compare the accuracy of PLMs using each low-rank approximation algorithm on MNLI, QQP, and SQuAD$_{1.1}$.}
	\label{tab:lra}
	\begin{tabular}{@{}lcccrrrr@{}}
		\toprule
		\multicolumn{1}{c}{\multirow{2}{*}{\textbf{Method}}} & \multirow{2}{*}{\textbf{Target$^\S$}} &  \multirow{2}{*}{\textbf{\begin{tabular}[c]{@{}c@{}} PLM$^{\text{\ding{168}}}$\end{tabular}}} &  \multirow{2}{*}{\textbf{\begin{tabular}[c]{@{}c@{}}Comp.\\ Rate\end{tabular}}} & \multicolumn{4}{c}{\textbf{Accuracy/F1 Difference$^\text{\dif}$ (\%p)}} \\
		\multicolumn{1}{c}{} & & & & \textbf{MNLI} & \textbf{QQP} & \textbf{SQD$^*$} &\textbf{Avg.} \\ \midrule
		Noach et al.~\cite{mf} & P & Bb & 40.73\% & -1.70 & 0.90 & - & -0.40 \\
		SVD~\cite{fwsvd} & P & Bb & 50.00\% & -1.90 & -0.50 & - & -1.20 \\
		FWSVD~\cite{fwsvd} & P & Bb & 50.00\% & -1.70 & -0.20 & - & -0.95 \\
		Wang et al.~\cite{exploring} & P & Bb & 85.70\% & 0.00 & 0.50 & - & 0.25 \\
		Wang et al.~\cite{exploring} & P & Bb & 97.91\% & -3.80 & -0.40 & - & -2.10 \\ \midrule
		LoRA~\cite{lora} & A & Rb & 0.24\%$^\P$ & -0.10 & -0.90 & - & -0.50 \\
		AdaLoRA~\cite{adalora} & A & Db & 0.17\%$^\P$ & 0.76 & -0.64& 0.90 & 0.34 \\ \bottomrule
		\multicolumn{8}{l}{$\P$: the ratio between the size of adaptors compared to the size of the original model}\\
		\multicolumn{8}{l}{
			\dif: (the accuracy (or F1 score) of a compressed model) - (the accuracy (or F1 score) of } \\
		\multicolumn{8}{l}{
			an uncompressed model), $\S$: P (original parameter), A (adaptor),
			*: $\text{SQuAD}_{1.1}$} \\
		\multicolumn{8}{l}{${\text{\ding{168}}}$: Bb (BERT-base), Rb (RoBERTa-base), Db (DeBERTaV3-base)}\\
	\end{tabular}
\end{table}

The first group of algorithms directly apply matrix (tensor) decomposition to the original parameters (P) in language models and directly reduce the memory and computational cost of language models~\cite{fwsvd,mf,exploring}.
Noach et al.~\cite{mf} apply LRA singular value decomposition (SVD) for each parameter matrix, and FWSVD~\cite{fwsvd} proposes to select singular vectors in SVD considering Fisher information which is directly related to the cross-entropy loss.
Wang et al.~\cite{exploring} achieve a remarkable compression rate over 97.91\% without significant accuracy degradation by exploiting tensor decomposition.
They generate a three-dimensional tensor by (1) concatenating all parameter matrices in each sublayer, and (2) vertically stacking the concatenated matrices.

The second group of algorithms called parameter-efficient fine-tuning (PEFT)~\cite{lora,adalora,dylora}
applies LRA to adaptors (A) to reduce the memory footprint of fine-tuning processes.
PEFT algorithms fix the original parameters in the language model and update only low-rank adaptors using gradient descent.
They merge low-rank adaptors into the original parameters after fine-tuning, thus the size of a language model does not change.
The value of PEFT algorithms lies in the significant reduction of the memory footprint during fine-tuning processes because of the reduced amount of gradients to save, and thus they are useful for fine-tuning billion-scale LLMs on limited computational resources.
LoRA~\cite{lora} is the first algorithm that introduces low-rank adaptors for fine-tuning language models.
However, finding the proper rank of each adaptor reflecting their importance is challenging. 
AdaLoRA~\cite{adalora} proposes an algorithm to find the proper rank for each adaptor by formulating the rank selection problem as a pruning problem: measuring the importance of each rank-one component and gradually pruning the least important ones.
PEFT algorithms contribute to improving the accuracy of pruning~\cite{pruner,loraprune} and quantization~\cite{NUQLoRA,loftq} algorithms for LLMs which requires memory-efficient fine-tuning processes to increase accuracy.
We explain more about PEFT algorithms with LoRA in Section~\ref{subsec:o6}. 

\subsection{Parameter Sharing}
\label{subsec:o4}

Parameter sharing is a technique that uses the same parameters in different parts of a model, and it promotes better generalization~\cite{tiedtrans} and fast convergence~\cite{pswang23} by reducing the number of parameters.
We categorize parameter sharing algorithms into sublayer sharing~\cite{unitrans, rsnmt, albert, takase21, subformer, peakd, lightformer} and matrix sharing~\cite{mqa, gqa} algorithms.

The sublayer sharing algorithms~\cite{unitrans, rsnmt, albert, takase21, subformer, peakd, lightformer} share the parameters in different sublayers.
There is an aggressive approach that shares all parameters in sublayers across a model~\cite{unitrans, rsnmt,albert}.
These algorithms significantly reduce the size of the model to be the same as that of a single-layer model, but they show an accuracy drop because of the reduced complexity of the compressed model.
Therefore, the latter studies~\cite{takase21, peakd} use group-wise sublayer sharing that shares the sublayer parameters within each group to increase the complexity of the compressed model.
Takase et al.~\cite{takase21} explore diverse grouping patterns of sublayers and Pea-KD~\cite{peakd} proposes a shuffled parameter sharing technique that shuffles query and key matrices when they share parameters.

The matrix sharing algorithms~\cite{mqa,gqa} share parameter matrices within a sublayer and they are widely used for compressing decoder-only Transformers.
The main idea of matrix sharing algorithms is to share key and value parameter matrices between attention heads within a sublayer to reduce the latency for loading key and value matrices in the decoding process.
MQA~\cite{mqa} shares key and value parameter matrices of all attention heads in each sublayer, whereas GQA~\cite{gqa} shares key and value parameter matrices within a group of attention heads to increase the accuracy of MQA.
GQA shows a better accuracy-latency trade-off than MQA, and GQA is adopted in LLaMA2-70B~\cite{llama2} which is one of the state-of-the-art LLMs.

\subsection{Efficient Architecture Design}
\label{subsec:o5}

Efficient architecture design is a technique that enhances the efficiency of language models by modifying their architecture.
We categorize those methods into two groups based on how they find better architectures: manual architecture design~\cite{sparsetransformer, reformer, linformer, mobilebert, funneltransformer} and neural architecture search (NAS)~\cite{evolvedtrans, adabert, hat, nasbert, YCS21, efficientbert}.

%
Manual architecture design methods~\cite{reformer,sparsetransformer,linformer,mobilebert,funneltransformer} manually find the better Transformer architecture to address inefficiencies of Transformer.
The major inefficiency of Transformer architecture lies in the expensive computational cost of a self-attention mechanism;
it is hard to process long texts with Transformer since self-attention mechanisms have $O(s^2)$ computational complexities for a length $s$ input sequence~\cite{transformer}.
Reformer~\cite{reformer} and Sparse Transformer~\cite{sparsetransformer} address this issue by reducing the number of tokens to attend for a self-attention mechanism.
Reformer reduces the computational cost of self-attention into $O(s\log s)$ by utilizing locality-sensitive hashing (LSH), and Sparse Transformer requires $O(s \sqrt{s})$ computations by factorizing attention operations.
Linformer~\cite{linformer} applies low-rank approximation to reduce the cost of self-attention and requires $O(sk)$ computations where $k$ is the reduced rank such that $k \ll s$.
On the other hand, MobileBERT~\cite{mobilebert} reduces the computational cost of Transformer by adopting an efficient bottleneck structure of MobileNet~\cite{mobilenet}, and Funnel-Transformer~\cite{funneltransformer} introduces pooling layers that reduce the sequence length of intermediate representations to reduce the computational cost of Transformer.


Neural architecture search (NAS) algorithms~\cite{hat,adabert,nasbert,efficientbert,YCS21} aim to find the most efficient and accurate architecture from a vast pool of possible architectures using deep learning algorithms.
They define the set of possible architectures called a search space and find candidate architectures based on their search strategy.
After that, they evaluate the selected candidate architectures following the predefined evaluation metric.
They gradually find better architecture candidates by reflecting on the evaluation results of previously selected architectures.
Architectures found by NAS algorithms show better accuracy-latency trade-off than manually designed architectures~\cite{efficientbert}, but the prohibitive cost for NAS limits their extension to large language models; NAS-BERT~\cite{nasbert} requires three days with 32 P40 GPUs for BERT-base which has only 110M parameters.

\subsection{Representative Algorithm: LoRA}
\label{subsec:o6}
Low-rank adaptation (LoRA)~\cite{lora} is an algorithm that fine-tunes pretrained language models for downstream tasks by introducing a small number of low-rank adaptors.
LoRA keeps the original parameters in the pretrained language model unchanged and updates only the low-rank adaptors to increase the efficiency of the adaptation processes.
%
The value of LoRA stems from the significantly reduced memory footprint during the fine-tuning process because of the reduced amount of gradients to memorize.
LoRA demonstrates its effectiveness by reducing the memory footprint from 1.2TB to 350GB for fine-tuning a GPT-3~\cite{gpt3} 175B model~\cite{lora}.

Consider a forward pass $\vh = \mW \vx$ that generates a $d_o$ dimensional output $\vh \in \bR^{d_o}$ using a fine-tuned weight matrix $\mW \in \bR^{d_o \times d_i}$ and a $d_i$ dimensional input $\vx \in \bR^{d_i}$.
LoRA separates $\mW$ into the sum of a pretrained weight matrix $\mW_0 \in \bR^{d_o \times d_i}$ and an incremental weight matrix $\Delta \mW \in \bR^{d_o \times d_i}$.
After that, LoRA approximates $\Delta \mW$ with the multiplication of two low-rank matrices $\mB\in\bR^{d_o\times r}$ and $\mA\in\bR^{r \times d_i}$ where $r$ is the rank of the low-rank matrices.
The modified forward pass for computing $\vh$ is represented in Equation~\eqref{eq:5_lora}.
\begin{equation} \label{eq:5_lora}
\vh = \mW \vx = \mW_0 \vx + \Delta \mW \vx \approx \mW_0 \vx + \mB \mA \vx
\end{equation}
After fine-tuning, $\mB$ and $\mA$ are merged into the original weight matrix $\mW_0$ so that the number of parameters of the language model is not changed.
One challenge of LoRA is to find a proper $r$ that maximizes the accuracy of the fine-tuned model since manually finding a proper $r$ by exmaning the accruacy of the fine-tuned models with all possible $r$ takes intractable time.
DyLoRA~\cite{dylora} fine-tunes LMs across a range of ranks instead of focusing on a single rank.
This method selects the rank that shows the best performance within the range of ranks.
LoRA and DyLoRA overlook the importance of individual weight matrix, and utilize the same rank for all weight matrices.
In contrast, AdaLoRA~\cite{adalora} tackles this limitation by dynamically assigning ranks based on the importance of each weight matrix.

The potential of LoRA stems from its compatibility and ease of integration with various compression algorithms, such as pruning~\cite{pruner, adalora} and quantization~\cite{NUQLoRA, UINT2.1}, as well as the parameter-efficient fine-tuning (PEFT) algorithms~\cite{adapter, prefix, bitfit, unipelt, s4}.
These algorithms utilize a fine-tuning process with LoRA to compensate for errors induced by compression to improve the performance of the compressed models.

\section{Discussion}
\label{sec:disc}

We summarize and discuss the survey results.
We compare categories of compression algorithms in Section~\ref{subsec:d1} and summarize our findings about two desired properties for successful compression algorithms in Section~\ref{subsec:d2}.
Finally, we propose promising research topics discovered through our survey in Section~\ref{subsec:d3}.

\subsection{Comparing Types of Compression Algorithms}
\label{subsec:d1}

We summarize the findings of our investigation on various language model compression algorithms.

The value of pruning algorithms~\cite{ziplm, cofi, kpruning,powerbert} is to provide an instant acceleration in inference on GPUs.
Pruning algorithms~\cite{ziplm,cofi} show the highest acceleration among compression algorithms with up to 15-fold faster performance than uncompressed models by pruning large granularities such as sublayers.
However, achieving such large granularity necessitates exhaustive training for error compensation and it limits the extension of its benefits to low-cost algorithms.

Quantization algorithms~\cite{UI-BERT,ULLM-QAT,NUSqueezeLLM,UOWQ} demonstrate high compression rates in both high-cost and low-cost scenarios.
In particular, quantization algorithms are applicable to the massive LLMs up to 175B, showing a maximum acceleration of 3.9 times faster inference speed than uncompressed models~\cite{UAWQ}.
However, it is observed that the bit-width for activations has not yet been reduced sufficiently, indicating the need for further research in this area~\cite{UZeroQuant,USmoothQuant,UBondarenko2023,URPTQ}.

Knowledge distillation (KD) algorithms~\cite{railkd,pkd} are essential for good performance in high-cost compression scenarios, since they improve the accuracy of compressed models~\cite{cofi,ziplm,dynabert}.
However, as low-cost compression algorithms for LLMs have become more important, their usage has diminished.
We need to find a way to efficiently apply knowledge distillation to low-cost compression scenarios to enhance the accuracy of compressed models.

Low-rank approximation (LRA) algorithms~\cite{exploring,fwsvd} have demonstrated the highest compression rates of parameters, up to 50 times, in high-cost compression scenarios.
However, LRA algorithms have not been widely used in low-cost compression scenarios for decoder-based LLMs because they require extensive retraining to retain the accuracy of the compressed models~\cite{exploring}.
Similar to KD algorithms, LRA algorithms require a breakthrough to achieve high accuracy without expensive retraining.

Notably, parameter-efficient fine-tuning (PEFT) algorithms~\cite{lora,adalora} contribute to improving the efficiency of the expensive fine-tuning process in LLMs.
PEFT algorithms enable more accurate fine-tuning results than local weight reconstruction in current low-cost compression scenarios since they directly reduce the objective function, such as cross-entropy.
We expect that various useful techniques previously utilized in high-cost compression scenarios such as KD and LRA can be applied to low-cost compression scenarios by improving PEFT algorithms to further reduce the cost of fine-tuning.

Lastly, extensive studies~\cite{sparsegpt, cofi, exploring,UBinaryBERT} have demonstrated that combining different compression algorithms results in high compression rates by mitigating diverse inefficiencies within the model.
However, there is a lack of research that unifies three or more compression algorithms at once.
Therefore, designing an algorithm to integrate diverse compression algorithms considering their characteristics is a promising approach to achieving extremely high compression rates without sacrificing accuracy.

\subsection{Two Desired Properties for Designing Successful Low-cost Compression}
\label{subsec:d2}

Our findings indicate that achieving successful language model compression requires satisfying the following two key properties.
First, it is crucial to directly incorporate a task-specific objective function instead of relying on local layer-wise reconstruction error.
Second, it is necessary to gradually compress the language model through a cost-efficient iterative process to minimize the amount of compression error in each iteration.

In high-cost compression scenarios, it is natural to perform gradient descent on the objective function.
However, many low-cost algorithms, especially those designed for LLMs, reduce their cost for compression by formulating local sublayer-wise reconstruction error as a least-square problem which takes a short time to solve.
Nevertheless, since the goal of compression is to minimize task-specific objective functions, such alternatives lead to suboptimal performance.
Prominent examples are FWSVD~\cite{fwsvd} and \kprune~\cite{kpruning}.
Although SVD theoretically guarantees minimal reconstruction error for matrices, FWSVD shows that incorporating SVD with gradient information regarding the task-specific objective function yields a significant performance improvement.
In \kprune, the authors observe that the pruned models designed to preserve only the knowledge regarding sublayer-wise reconstruction errors show inferior accuracy. They achieve a significant performance improvement when they make the model additionally preserve the knowledge regarding a task-specific objective function.
Modifying the objective function as a task-specific objective function is a promising approach to enhance the performance of the algorithms~\cite{sparsegpt,wanda,UOPTQ,URPTQ} that utilize proxy objective functions, such as a sublayer-wise reconstruction error.

Compression algorithms consist of two stages: modifying the representation of weights to make them computation or memory efficient, and compensating for compression errors induced by the modification.
In most PLM compression scenarios, there is no access to a large corpus used for pertaining PLMs, but only relatively small task-specific datasets or even smaller calibration datasets are available.
Therefore, PLM compression algorithms should preserve useful features learned from the large corpus since they cannot be reconstructed using a small dataset.
High-cost compression algorithms successfully preserve the useful features since they gradually transform weights and compensate for compression errors.
However, low-cost compression algorithms tend to transform all weights at once and lose the useful features of the PLM.
Though they attempt to recover the useful features of the PLM through weight-tuning processes, they fail to recover the lost features.
For example, LLM-pruner~\cite{pruner} fails to restore the PLM's performance under a low compression rate of 20\% even though it fine-tunes the compressed model with LoRA~\cite{lora}.
One notable approach is Kprune~\cite{kpruning}.
\kprune addresses the accuracy degradation problem of low-cost compression algorithms by employing an iterative compression process consisting of mild pruning and error compensation.
As a result, Kprune achieves significant performance improvement compared to other low-cost compression algorithms~\cite{KKMHKG22,kcm} while maintaining efficiency.
The benefit of iterative compression algorithms is also reported in numerous works~\cite{pruning,lottery,SSM20} for other types of neural networks, therefore it is a promising approach to design a cost-effective and iterative low-cost algorithm to achieve high accuracy.

\subsection{Promising Research Topics}
\label{subsec:d3}

We introduce promising research areas discovered through our survey as follows.
We focus on research topics related to low-cost compression algorithms considering their heightened significance for LLMs.

\begin{itemize*}
	\item \textbf{Low-cost iterative algorithms.}
	How can we efficiently apply iterative compression algorithms to LLMs to improve the accuracy of compressed models?
	\item \textbf{Directly optimizing the objective.}
	How can we directly reduce the target objection function rather than proxy objective functions such as sublayer-wise reconstruction errors in LLMs?
	\item \textbf{Integrating PEFT with high-cost algorithms.}
	How can we leverage PEFT algorithms to replace expensive retraining processes of accurate high-cost algorithms, such as knowledge distillation or low-rank approximation algorithms?
	\item \textbf{Accurate pruning algorithms for LLMs.}
	Pruning algorithms show severe accuracy degradation for LLMs even under low compression rates. What aspects of LLMs degrade the performance of pruning algorithms, and how can we improve the accuracy of them?
	\item \textbf{Activation quantization for LLMs.}
	Activation quantization is a necessary technique to increase the inference speed of quantized models, but quantization algorithms fail to retain the accuracy of compressed models whose bit-width is below 8.
	How can we quantize the activations of LLMs with low-bits maintaining their accuracy?
	\item \textbf{Unifying Compression algorithms.}
	Each type of compression algorithm addresses different kinds of inefficiencies and they are able to be unified to achieve higher compression rates.
	How can we unify diverse compression algorithms to achieve extremely high compression rates?
\end{itemize*}

\section{Conclusion}
\label{sec:conclusion}
In this paper, we perform an extensive survey of language model compression algorithms including pruning, quantization, knowledge distillation, low-rank approximation, parameter sharing, and efficient architecture design algorithms.
We cover compression algorithms for both encoder-only and decoder-only Transformers.
We elaborate on the details of background knowledge and three representative algorithms to promote a better understanding of readers.
We discuss the value of each compression algorithm type, and two desired properties for designing successful low-cost compression algorithms.
Finally, we introduce promising research topics focusing on cost-effective compression algorithms for LLMs which are the most important language models.

\begin{acks}
This work was supported by Youlchon Foundation. This work was also supported by Institute of Information \& communications Technology Planning \& Evaluation(IITP) grant funded by the Korea government(MSIT) [No.2020-0-00894, Flexible and Efficient Model Compression Method for Various Applications and Environments], [No.2021-0-01343, Artificial Intelligence Graduate School Program (Seoul National University)], and [NO.2021-0-02068, Artificial Intelligence Innovation Hub (Artificial Intelligence Institute, Seoul National University)]. The Institute of Engineering Research at Seoul National University provided research facilities for this work. The ICT at Seoul National University provides research facilities for this study. U Kang is the corresponding author.
\end{acks}
\bibliographystyle{ACM-Reference-Format}
\bibliography{ref_general,ref_pruning,ref_quant,ref_other}

\end{document}